\documentclass[preprint]{cas-dc}

\usepackage[numbers]{natbib}
\usepackage{amsmath,amssymb,amsfonts}
\usepackage{algorithmic}
\usepackage{graphicx}
\usepackage{textcomp}
\usepackage{diagbox}
\usepackage{float}
\usepackage{url}
\usepackage{amsfonts}
\usepackage{graphicx}
\usepackage{subfig}
\usepackage{rotating}
\graphicspath{{figures/}}
\usepackage{multirow}
\usepackage{hhline}
\usepackage{amsmath}
\usepackage{adjustbox}
\usepackage{caption}
\usepackage{fancyhdr}
\usepackage{rotating}
\usepackage{hyperref}
\usepackage{bm}
\usepackage{makecell}
\usepackage{mleftright}
\usepackage{comment}

\DeclareMathOperator*{\argmax}{arg\,max}

\def\tsc#1{\csdef{#1}{\textsc{\lowercase{#1}}\xspace}}
\tsc{WGM}
\tsc{QE}

\begin{document}
\let\WriteBookmarks\relax
\def\floatpagepagefraction{1}
\def\textpagefraction{.001}

\shorttitle{Neuromorphic vision based control}    

\shortauthors{Ayyad el al.}  

\title[mode = title]{Neuromorphic Vision Based Control for the Precise Positioning of Robotic Drilling Systems}

  \tnotemark[1]

  \tnotetext[1]{This work was performed as part of the Aerospace Research and Innovation Center (ARIC), which is funded by STRATA Manufacturing PJSC (a Mubadala company), Khalifa University of Science and Technology, and RC1-2018-KUCARS (Khalifa University Center for Autonomous Robotic Systems).}

  \author[1,2]{Abdulla Ayyad}[type=author,
    orcid=0000-0002-3006-2320]
  \cormark[1] 
  \fnmark[1] 
  \ead{abdulla.ayyad@ku.ac.ae} 

  \credit{Conceptualization, Methodology, Software, Investigation, Writing}

  \author[1,2]{Mohamad Halwani}

  \credit{Software, Experimentation, Data collection, Writing}
  
  \author[3]{Dewald Swart}

  \credit{Conceptualization, End-effector Design, Software}
  
  \author[2]{Rajkumar Muthusamy}
  
  \credit{Methodology, Technical Advising}
  
  \author[4]{Fahad Almaskari}
  
  \credit{Supervision, Review and Editing}
  
  \author[1,2,4]{Yahya Zweiri}

  \credit{Project management, Funding acquisition, Review and Editing}

  \address[1]{Aerospace Research and Innovation Center (ARIC), Khalifa University of Science and Technology, Abu Dhabi, United Arab Emirates}
  \address[2]{Khalifa University Center for Autonomous Robotic Systems (KUCARS), Khalifa University of Science and Technology, Abu Dhabi, United Arab Emirates}
  \address[3]{Research and Development, Strata Manufacturing PJSC, Al Ain, United Arab Emirates}
  \address[4]{Department of Aerospace Engineering, Khalifa University, Abu Dhabi, United Arab Emirates}

  \cortext[cor1]{Corresponding author}

 \begin{abstract}[S U M M A R Y]
  The manufacturing industry is currently witnessing a paradigm shift with the unprecedented adoption of industrial robots, and machine vision is a key perception technology that enables these robots to perform precise operations in unstructured environments. However, the sensitivity of conventional vision sensors to lighting conditions and high-speed motion sets a limitation on the reliability and work-rate of production lines. Neuromorphic vision is a recent technology with the potential to address the challenges of conventional vision with its high temporal resolution, low latency, and wide dynamic range. In this paper and for the first time, we propose a novel neuromorphic vision based controller for faster and more reliable machining operations, and present a complete robotic system capable of performing drilling tasks with sub-millimeter accuracy. Our proposed system localizes the target workpiece in 3D using two perception stages that we developed specifically for the asynchronous output of neuromorphic cameras. The first stage performs multi-view reconstruction for an initial estimate of the workpiece's pose, and the second stage refines this estimate for a local region of the workpiece using circular hole detection. The robot then precisely positions the drilling end-effector and drills the target holes on the workpiece using a combined position-based and image-based visual servoing approach. The proposed solution is validated experimentally for drilling nutplate holes on workpieces placed arbitrarily in an unstructured environment with uncontrolled lighting. Experimental results prove the effectiveness of our solution with an average positional errors of less than 0.1 mm, and demonstrate that the use of neuromorphic vision overcomes the lighting and speed limitations of conventional cameras. The findings of this paper identify neuromorphic vision as a promising technology that can expedite and robustify robotic manufacturing processes in line with the requirements of the fourth industrial revolution.
 \end{abstract}

 \maketitle

\section{Introduction}
\label{intro}

The fourth industrial revolution shows significant emphasis on the automation of high-precision cyber-physical manufacturing and machining processes. Automating such processes offer numerous advantages in terms of performance, productivity, efficiency, and safety; as manual operation is often associated with structural damage, risk of rework, and health hazards \cite{Liang2010, Eguti2014, Zhu2014, Mei2021}. Among other processes, drilling has been studied extensively by academics and practitioners  due to their widespread use in various manufacturing activities, especially in the automotive and aerospace industries \cite{Karim2013, Liang2010, Zhan2012, Zhang2018a}. High-precision drilling is essential for these industries, as the quality of drilling is highly correlated with the performance and fatigue life of the machined structures \cite{Mei2021, Liu2016, Liang2010, Sun2019, Chen2018}.

Traditionally, the automation of drilling and similar machining processes has been highly dependant on Computer Numerical Control (CNC) equipment for their high-precision and repeatability. However, CNC equipment are limited in functionality and workspace, and require substantial investment in both equipment and infrastructure \cite{Ji2019, Liang2010}. In recent years, industrial robots have been rising as a promising alternative for CNC equipment in machining applications due to their cost efficiency, their wide range of functionality, their ability to operate on large workspace volumes and their capability to adapt to variations in the environment and workpiece positioning \cite{Chen2013, Eguti2014, Karim2013, Ji2019, Devlieg2011}. Despite several successful examples of utilizing robots in industrial machining applications, repeatability remains the main challenge in robotic machining; where errors originate either from the relatively low stiffness of robot joints \cite{Eguti2014, Sun2019, Olsson2010, Zhang2020} or the imperfect positioning and localization of a workpiece relative to the robot \cite{Perez2016, Bone2003}. These errors can be undermined by the use of real time guidance and closed-loop control based on sensory feedback and metrology systems \cite{Huang2018, Eguti2014, Ji2019, Perez2016}. Several works in the literature have adopted such approaches for precise position \cite{Jia2018, Wang2020}, orientation \cite{Rao2020, Rao2018}, and force \cite{Olsson2010, Rosa2017} control in a robotic machining paradigm. 

Machine vision is amongst the most utilized perception technologies that enable the closed-loop control of robots due to their maturity, availability, and relatively low cost \cite{Perez2016}. In \cite{Zhu2014}, a 2D vision system was used to enhance the drilling positional accuracy through the detection and localization of several reference holes in the workpiece; achieving a position accuracy of 0.1 mm. Similarly, the work in \cite{Mei2021} utilizes feedback from an eye-in-hand camera and uses template matching to localize reference holes for a combined drilling and riveting process, reducing positioning errors to ~0.05 mm. \cite{Frommknecht2017} proposed combining the 2D camera detection with laser distance sensors to localize reference holes in 3D, and reported an accuracy of 0.3 mm. Similar approaches for the positioning of a drilling tool are reported in \cite{Zhan2012, Liu2016}; with variations in the underlying perception and reference hole detection algorithms. Several works in the literature focus exclusively on the robust detection of circular holes through contour refinement and model fitting due to its direct impact on the precision of the drilling process \cite{Lou2020, Mei2015, Xia2020}. These concepts of vision-based feature detection and workpiece localization are also widely adopted in other various manufacturing tasks \cite{Ji2019, Yu2019}. For instance, \cite{Jiang2021} proposed a visual guidance system for robotic a peg-in-hole application consisting of four cameras: two in an eye-to-hand configuration for the localization of the robotic tool, while the others are in an eye-in-hand configuration and are used for alignment of the tool with reference holes. A multi-view approach was presented in \cite{Yang2020} for the localization of target objects in a pick-and-place framework with sub-millimeter level accuracy. The versatility of vision systems have also enabled other uses in navigation, guidance, and calibration systems of mobile industrial robots \cite{Zhao2019, Mei2015a, Guo2017}.  

All of the aforementioned robotic manufacturing approaches utilize conventional frame-based cameras, which suffer from latency, motion blur, low dynamic range and poor perception at low-light condition \cite{wang2020eye, Corke2000}. Frame-based cameras output intensity images based on a time integration of incident illumination over a fixed exposure period. This integral action introduces latency in perception, and causes blurring in the image when considerable relative motion exists; especially with larger exposure periods \cite{1996corke, Shin2019}. On the other hand, short exposure times greatly degrade image clarity in reduced lighting conditions, and require larger apertures which leads to a narrow depth of field \cite{1996corke, Pieters2013}. These shortcomings of frame-based cameras impose constraints on robot operational speeds, workspace volumes, and ambient lighting conditions; which affect the robustness and productivity of robotic manufacturing processes. Relevant work in the literature attempt to mitigate these problems with conventional cameras by adding additional supporting sensors \cite{SANTOS2022}; which increases the complexity and cost of the system. 

The recent neuromorphic vision sensor (also known as event-based camera) has the potential to address the challenges of conventional machine vision. The pixels of a neuromorphic camera operate independently and respond asynchronously to variations in incident illumination in continuous time \cite{Lichtsteiner2008,indiveri2000neuromorphic}, providing low-latency, high dynamic range, and computationally efficient perception \cite{gallego2019event, Liu2019neuromorphic}. Neuromorphic cameras do not suffer from motion blur, and are robust to varying lighting conditions; making them an appealing choice for a wide variety of applications such as: autonomous driving \cite{Dong2020ADAS, Guang2020ADAS}, Unmanned Aerial Vehicle Control \cite{Hay2021}, object recognition and tracking \cite{Garrick2015recognition, Hongmin2010tracking}, localization and mapping \cite{Kim2016slam, visal2018slam}, and tactile sensing \cite{rajkumar2020slip, huang2020neuromorphic, Rigi2018}. In our recent work \cite{rajkumar2021}, we have demonstrated the advantages of neuromorphic cameras over their conventional frame-based counterparts for high-speed and uncontrolled lighting operation in a robotic pick-and-place framework. However, the low-resolution of the neuromorphic camera, the assumption of known depth, and the act-to-perceive nature of the event camera resulted in positional errors of up to 2 cm. 

In this paper, we develop and employ a two-stage neuromorphic vision-based controller to perform a robotic drilling task with sub-millimeter level accuracy. The first stage localizes the target workpiece in 6DoF using a multi-view 3D reconstruction approach and Position Based Visual Servoing (PBVS). The second control stage applies Image Based Visual Servoing (IBVS) to compensate for positional errors using and a set of reference holes. Using both control stages, the robot performs peg-in-hole to insert a clamp mandrel (or split-pin) in the reference hole with less than 0.2 mm clearance. The clamp mandrel holds the robotic tool in place, and the robot drills nutplate installation holes on both side of each reference hole. The capabilities of the neuromorphic camera enables higher-speed operation and robustness to changes in ambient lighting. A video demonstrating the presented robotic drilling solution can be accessed through this link: \url{https://drive.google.com/file/d/1q9QwPvkd7ZcEBcGMIxIVy2r_iRfTKCSe/view?usp=sharing}
\cite{paper_video}.

The contributions of this paper can be summarized as below:
\begin{enumerate}
    \item   For the first time, we present a neuromorphic vision-based control approach for robotic machining applications. The proposed method utilizes the feedback of a neuromorphic camera to precisely align a drilling tool with the target workpiece, and demonstrates advantages over conventional vision-based solutions in high speed operation and uncontrolled lighting conditions. 
    
    \item   We devise an event-based multi-view 3D reconstruction method for the 6DoF localization of workpiece in the environment. This method matches events generated from different poses of the neuromorphic camera and solves for the 3D location of workpiece features using the Direct Linear Transformation (DLT).   
    
    \item   We develop a novel event-based approach for the detection and tracking of circular objects in the scene. This approach applies an event-based variant of the circle hough transform in a bayesian framework to detect and track the location of reference holes in the workpiece. 
    
    
    \item   We perform rigorous experimentation to test the precision and performance of the proposed methods. Experimental results validates the use of neuromorphic vision for robotic machining applications with positional accuracy of ~0.1 mm, and prove that our approach overcomes the speed and lighting challenges in conventional vision-based robotic machining approaches.
\end{enumerate}

The remainder of this paper is organized as follows. Section \ref{sec_setup} outlines the setup and configuration of the proposed robotic drilling system. Section \ref{sec_neuro_vision} describes the working principle and  functional advantages of neuromorphic cameras. Section \ref{sec_multi_view} explains the event-based multi-view 3D reconstruction and workpiece localization algorithm. Section \ref{sec_hole_det} introduces the event-based circular hole detection and tracking pipeline. Section \ref{sec_control} presents the two-stage vision-based controller employing both PBVS and IBVS. Finally, Section \ref{sec_exp} demonstrates both quantitative and qualitative experimental evaluation of the presented approach, which confirm the advantages of using neuromorphic vision for precise robotic processes.

\begin{figure*}[!t]
 \centering
      \subfloat[]{\label{fig:sf1}
      \includegraphics[width=0.6\textwidth, height=0.3\textwidth]{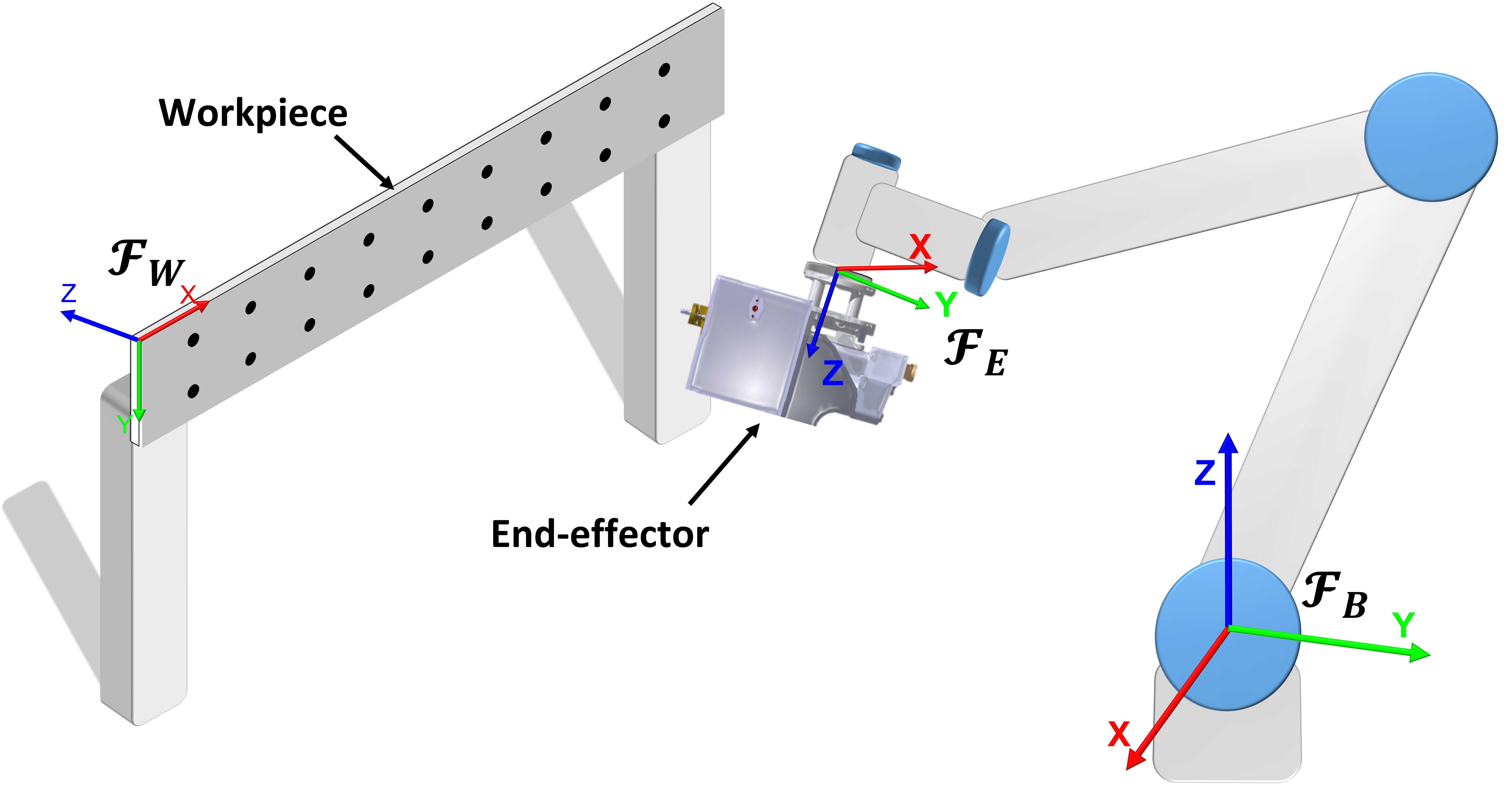}}  \qquad
      \subfloat[]{\label{fig:sf1}
      \includegraphics[width=0.33\textwidth]{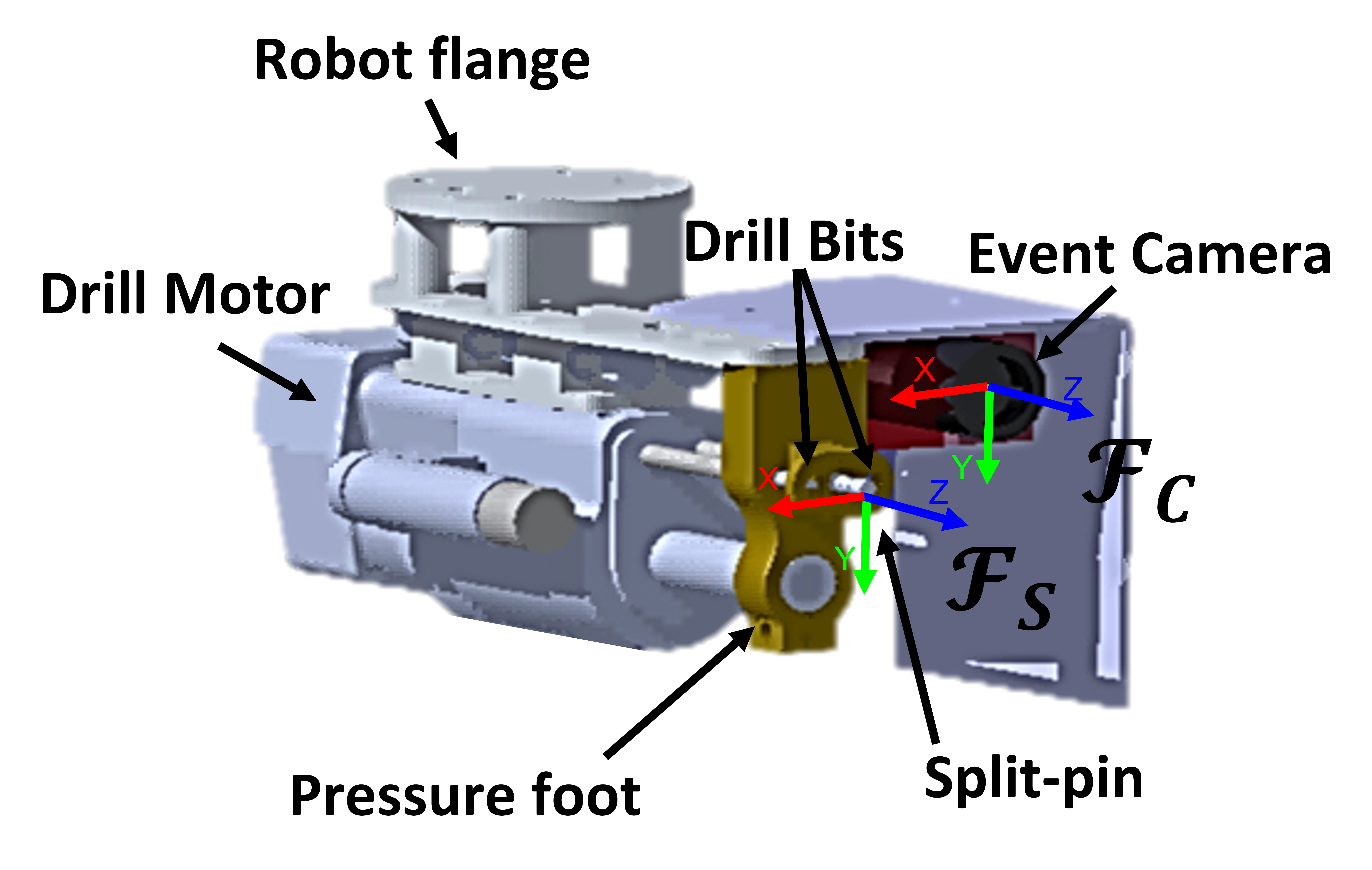}}  \qquad
\caption{The robotic nutplate hole drilling setup. (a) A workpiece with a set of reference holes is present within the industrial robot's workspace. (b) The nutplate hole drilling end-effector with a neuromorphic camera.}
\label{fig_drilling_setup}
\vspace{-1em}
\end{figure*}

\section{Robotic Drilling Setup}
\label{sec_setup}

The overall configuration of the robotic nutplate hole drilling system can be seen in Figure \ref{fig_drilling_setup}. The system consists of an industrial robot with an end-effector comprising a drill motor and a neuromorphic vision sensor for perception and guidance. The robotic system drills nutplate installation holes on a workpiece that includes a set of reference holes. We define the following frames of reference that are used throughout this paper for robot guidance and control:
\begin{itemize}
    \item \(\mathcal{F}_B\): The robot base coordinate frame.
    \item \(\mathcal{F}_E\): The robot end-effector coordinate frame.
    \item \(\mathcal{F}_C\): The vision sensor coordinate frame.
    \item \(\mathcal{F}_S\): The split-pin coordinate frame.
    \item \(\mathcal{F}_W\): The workpiece coordinate frame. 
    \item \(\mathcal{F}_{h^i}\): The coordinate frame of the i'th reference hole.
\end{itemize}

We denote the rotation matrix that maps from a source frame \(\mathcal{F}_S\) to a target frame \(\mathcal{F}_T\) by \(_{T}R_{S} \in \mathbb{R}^{3\times3}\). The position of point \(b\) relative to point \(a\) described in coordinate frame \(\mathcal{F}_T\) is given by \(^{T} _{a}\vec{P}_{b} \in \mathbb{R}^3\). As such, we define the affine transformation matrix \(_{T}T_{S} \in \mathbb{R}^{4\times4}\) as follows:

\begin{equation}
    \label{eq_affine_transformation}
    _{T}T_{S} = \begin{bmatrix}
    _{T}R_{S} & ^{T}_{T}\vec{P}_{s} \\
    \mathbf{0}^T & 1
    \end{bmatrix}
\end{equation}

For the remainder of this paper, we consider the transformation from \(\mathcal{F}_B\) to \(\mathcal{F}_E\) to be known by solving the robot's forward kinematics:
\begin{equation}
    \label{eq_forward_kinematric}
    _{B}T_{E} = g(\theta),      \theta \in \mathbb{C}
\end{equation}
where $g(\theta)$ is a nonlinear function representing the robot's kinematics, $\theta$ are the observed robot joint angles, and $\mathbb{C}$ is the robot's configuration space. Furthermore, $_{E}T_{C}$ and $_{C}T_{S}$ are constants, and can be found using a geometrical calibration procedure as described in \cite{Dornaika1998}. Therefore, $_{B}T_{C}$ and $_{B}T_{S}$ can be easily computed by combining $_{B}T_{E}$ and the calibrated transformations.

Similarly, the robot's forward kinematics are used to find the end-effector's twist vector \(\vec{\mathcal{V}}_E \in \mathbb{R}^6\) combining linear and angular velocity components, as follows:
\begin{equation}
    \vec{\mathcal{V}}_E = J(\theta)^{\dagger} \dot{\theta}
    \label{eq_vel_forward_kinematics}
\end{equation}
\begin{equation}
J(\theta) \triangleq \dfrac{\partial g}{\partial \theta} \in \mathbb{R}^{6 \times N_j}
 \label{jacobian}
\end{equation}
where \(J(\theta)\) is the Jacobian matrix, \(N_j\) is the number of robot joints, and \(\dagger\) denotes the Moore-Penrose inverse computed using Singular Value Decompision. Additionally, the camera's twist vector \(\vec{\mathcal{V}}_C\) can be calculated from \(\vec{\mathcal{V}}_E\) using the adjoint representation of $_{C}T_{E}$, denoted by $_{C}[AD_T]_{E}$, as follows:
\begin{equation}
 \vec{\mathcal{V}}_C = _{C}[AD_T]_{E} \vec{\mathcal{V}}_E
 \label{eq_camera_twist}
\end{equation}
\begin{equation}
 _{C}[AD_T]_{E} = \begin{bmatrix}
 _{C}R_{E} & & & [^{C}_{C}\vec{P}_{E}]_x \hspace{0.1cm }_{C}R_{E} \\
 \mathbf{0}^T & & & _{C}R_{E}
 \end{bmatrix}
 \label{eq_adjoint_representation}
\end{equation}
where \([^{C}_{C}\vec{P}_{E}]_x\) denotes the matrix representation of the cross product for the vector \(^{C}_{C}\vec{P}_{E}\).

In order to perform the drilling operation, the robot requires knowledge of $_{B}T_{h^i}$. We solve for this transformation in two stages. First, a multi-view 3D reconstruction approach provides an initial estimate of $_{B}T_{h^i}$ for all the workpiece holes as described in section \ref{sec_multi_view}. Then, for each hole, $_{B}T_{h^i}$ is refined to sub-millimeter accuracy using the circular hole detection and tracking approach presented in section \ref{sec_hole_det}. Following these two perception stages, robot control is performed by two subsequent methods: PBVS and IBVS; that align \(\mathcal{F}_S\) with \(\mathcal{F}_{h^i}\) and drills the required holes in the workpiece, which is explained in detail in section \ref{sec_control}.

\section{Neuromorphic Vision Sensor}
\label{sec_neuro_vision}

\begin{figure}[!t]
 \centering
\includegraphics[width=\linewidth]{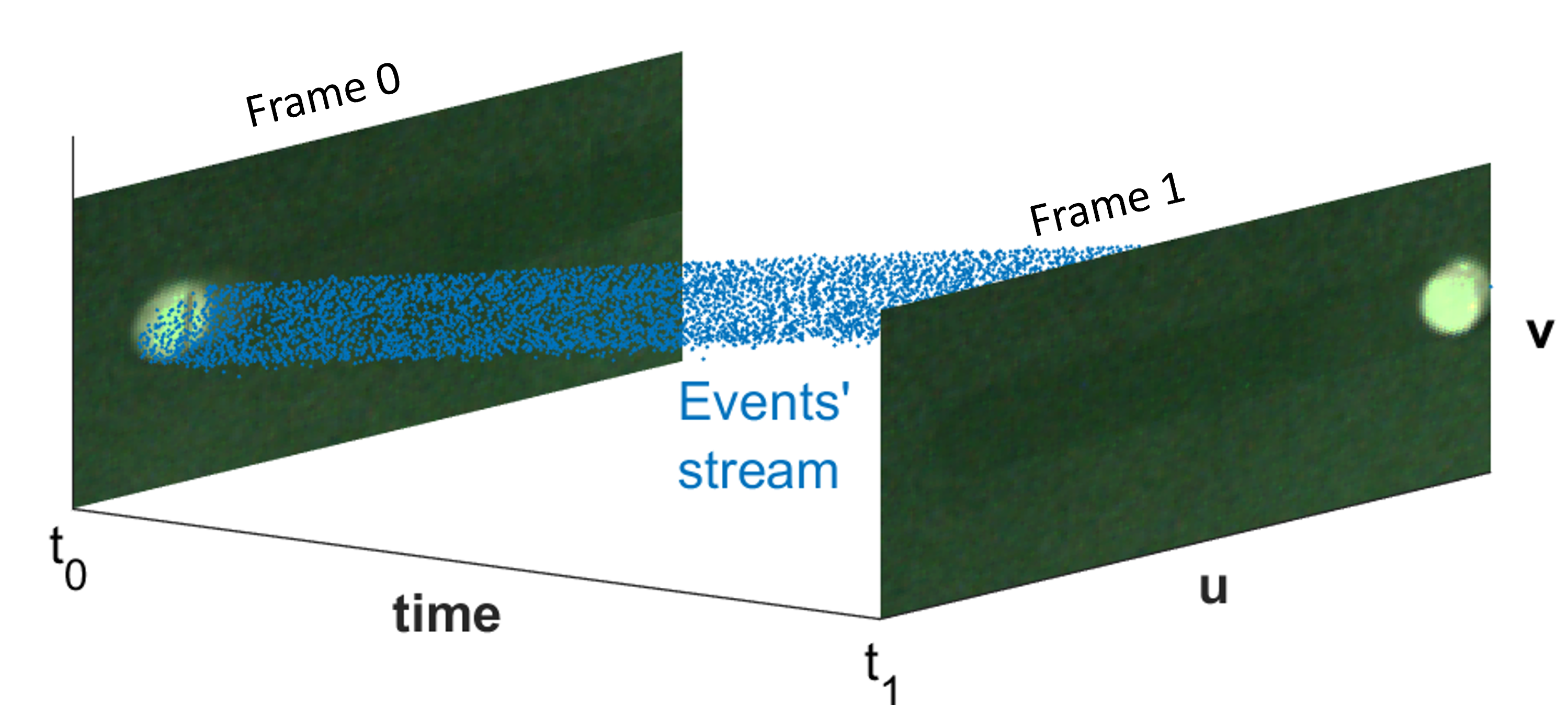}
\caption{Visualization of the output of conventional frame-based imagers and the event stream output of neuromorphic cameras.}
\label{fig_events_Vs_frames}
\vspace{-1em}
\end{figure}

The Neuromorphic vision sensor, often referred to as `event camera', decodes illumination changes in the visual scene as a steam of events \(e_{k}=<u_k, v_k, t_k, p_k>\), where \((u_k, v_k)\) represent the pixel coordinates of the change, \(t_k\) is the event's timestamp, and \(p_k\) is the illumination change polarity (either 1 or -1). Unlike conventional frame-based imagers, neuromorphic vision sensors do not operate on a fixed sampling rate; instead, pixels operate asynchronously and respond to logarithmic illumination changes with microsecond resolution. Figure. \ref{fig_events_Vs_frames} visualizes the differences between neuromorphic event-based cameras and conventional imagers. Let \(I(u, v, t)\) denote the illumination intensity at pixel \((u,v)\) and time \(t\), an event \(e=<u, v, t, p\) is triggered as soon as the following condition is met:
\begin{equation}
    log I(u, v, t) - log I(u, v, t-\Delta t) = p C
    \label{eq_event_condition}
\end{equation}
where \(C\) is the logarithmic illumination change threshold, and \(\Delta t\) is the time since the last triggered event at \(u, v\).

The working principle of neuromorphic cameras provides substantial advantages over conventional imaging sensors. For instance, neuromorphic cameras offer high temporal resolution and an exceptionally low-latency in the order of microseconds; meaning that these sensors do not suffer from motion blur and guarantee timely perception of changes in the scene \cite{gallego2019event}. Additionally, since independent pixels are self-sampled, neuromorphic cameras have a wide dynamic range \((>120 dB)\) \cite{Liu2019neuromorphic}, and are not impeded by the exposure timing complications that arise in frame-based cameras. This enables neuromorphic cameras to offer robust perception across a variety of lighting conditions, including extremely low-light cases. Another practically valuable feature that result of the aforementioned capabilities of neuromorphic vision is the ability to perceive under very small aperture, leading to a substantially wide depth of field. In applications such as ours where the camera is expected to acquire information across a varied depth, this feature can alleviate the need of an autofocus system, which often requires additional hardware \cite{wang2021} and induces uncertainty in the camera projection model \cite{viala2021}. Other advantages of neuromoorphic vision include low power consumption and reduction in signal redundancy as only informative data is transmitted in the form of events. Despite the capabilities of neuromorphic vision, the fundamentally different output of these cameras require novel computer vision algorithms and processing techniques than those developed for conventional frame-based imaging. 

It must be noted that neurmorphic cameras use identical optics as conventional cameras. As such, the standard pinhole model can still describe the projection properties of neuromorphic cameras. Following the pinhole model, the mapping between a point in three dimensional space \(^{B} _{B}\vec{P}_{a} = \begin{bmatrix} x_a & y_a & z_a \end{bmatrix}^T\) described in coordinate frame \(\mathcal{F}_B\), and its projection on the image plane \((u_a, v_a)\) can be expressed in homogeneous coordinates as follows:

\begin{equation}
    \begin{bmatrix}u_a,v_a,1\end{bmatrix}^T \sim \bf{K} \hspace{1em} _{C}T_{B} \hspace{1em}
    \begin{bmatrix} x_a & y_a & z_a & 1 \end{bmatrix}^T
 \label{eq_pinhole}
\end{equation}
where \(\sim\) indicates equality up to an unknown scalar multiplication, and \(\bf{K}\) denotes the camera intrinsic matrix. In this paper, we consider \((u, v)\) to be the pixel coordinates post rectification for tangential and radial distortions.

\section{Neuromorphic Event-based Multi-View Workpiece localization}
\label{sec_multi_view}

This section presents the event-based multi-view 3D reconstruction method used for the 6-Dof localization of the workpiece. As described in section \ref{sec_dlt}, we utilize the camera's projective geometry and the Direct Linear Transformation (DLT) to solve for the location of each reference hole in the workpiece using their corresponding stream of events from multiple camera viewpoints. We establish correspondences between the asynchronous events and reference holes using the space-sweep approach described in \ref{sec_space_sweep}. Finally, we use model fitting to determine the correct orientation of the workpiece.

\subsection{The Direct Linear Transformation}
\label{sec_dlt}

As the camera moves in the environments, a stream of events will be generated corresponding to each reference hole in the workpiece. The objective of the direct linear transformation is to determine the holes' position from their corresponding events generated at \(N\) different time-steps. At a given time-step \(k\) and camera pose \(_{C}T_{B}^k\), the relationship between the position of the i'th reference hole \(^{B}_{B}\vec{P}_{h^i}\) and its corresponding event in homogenous coordinates \(eh^i_k = [u^i_k, v^i_k, 1]\) can be described using the pinhole model in eq. \eqref{eq_pinhole}. By multiplying both sides of eq. \eqref{eq_pinhole} by \([eh^i_k]_x\), which is the matrix representation of the cross product for vector \(eh^i_k\), we obtain the following expression:

\begin{equation}
    [eh^i_k]_x \hspace{0.3em} eh^i_k = 
    \mathbf{0} = 
    \hspace{0.3em} [eh^i_k]_x \hspace{0.3em} \mathbf{K} \hspace{0.3em} _{C}T_{B}^k \hspace{0.3em} \begin{bmatrix}
    ^{B}_{B}\vec{P}_{h^i} \\ 1
    \end{bmatrix}
 \label{eq_dlt_step1}
\end{equation}

We define a matrix \(A_i \in \mathbb{R}^{3N\times4}\) that encapsulates the right side of eq. \eqref{eq_dlt_step1} across \(N\) observations of the i'th reference hole as follows:

\begin{equation}
    A_i = \begin{bmatrix}
    \hspace{0.3em} [eh^i_1]_x \hspace{0.3em} \mathbf{K} \hspace{0.3em} _{C}T_{B}^1 \\
    \hspace{0.3em} [eh^i_2]_x \hspace{0.3em} \mathbf{K} \hspace{0.3em} _{C}T_{B}^2 \\
    \vdots \\
    \hspace{0.3em} [eh^i_N]_x \hspace{0.3em} \mathbf{K} \hspace{0.3em} _{C}T_{B}^N \\
    \end{bmatrix}
 \label{eq_dlt_step2}
\end{equation}

From eqs. \eqref{eq_dlt_step1} and \eqref{eq_dlt_step2}, it is evident that the following expression holds:
\begin{equation}
    \mathbf{0} = A_i \hspace{0.3em} \begin{bmatrix}
    ^{B}_{B}\vec{P}_{h^i} \\ 1
    \end{bmatrix}
 \label{eq_dlt_step3}
\end{equation}

Eq. \eqref{eq_dlt_step3} is used to obtain a valid solution for \(^{B}_{B}\vec{P}_{h^i}\) as a least square problem, which can be efficiently solved using Singular Value Decomposition.

\subsection{The Space-Sweep Method}
\label{sec_space_sweep}

Reconstruction using the Direct Linear Transformation described in section \ref{sec_dlt} requires accurate correspondence between features in the environment (reference holes) and the events generated from different camera views. For this objective, we use the space-sweep method first introduced in \cite{collins1996} and adapted for neuromorphic cameras in \cite{emvs2018}. 
This approach utilizes a descritized representation of the volume of interest, denoted by \(\bar{D} \in \mathbb{R}^{w \times h \times N_z}\) where \(w\), \(h\), and \(N_z\) indicate the width, heigh, and depth of \(\bar{D}\). Each generated event is then back-projected as a ray passing through \(\bar{D}\) using the camera's pose and projection model. A Disparity Space Image (DSI) is defined that records the density of rays passing through each voxel of \(\bar{D}\). Local maxima are then extracted from the DSI, and the rays passing through high-density voxels are clustered together and are considered to correspond to the same feature in 3D space. Consequently, since each ray is defined by a camera pose and an event, the clustering of events and camera poses is inferred directly from these rays, and the DLT can hence be applied for each cluster independently. Figure \ref{fig_EMVS} visualizes the principal of the space-sweep method, and illustrates the different steps involved in this process. We refer interested readers to the original manuscripts in \cite{collins1996, emvs2018} for the details on the computationally efficient implementation of the Space-Sweep method.

\begin{figure}[t!]
 \centering
      {\label{fig_EMVS}
      \includegraphics[width=\linewidth]
      {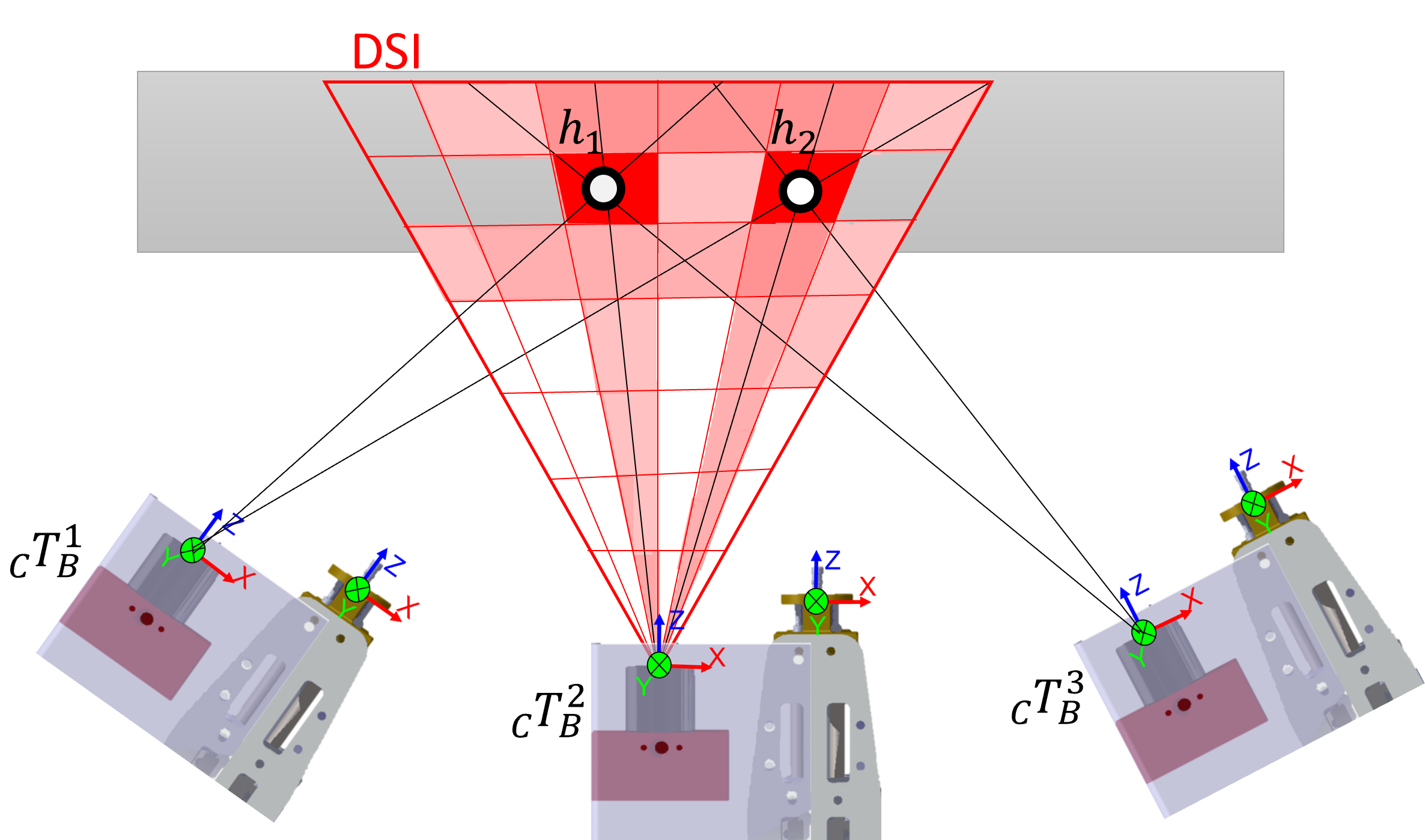}}  \qquad
\caption{The Event-Based space-sweep step. Events generated at different camera viewpoints are back-projected and are used for the voting process in the DSI. The rays passing through each of the dense voxels of the DSI are grouped together, and are then used to solve for the 3D location of the reference holes through the Direct Linear Transform.}
\label{fig_EMVS}
\vspace{-1em}
\end{figure}

After estimating the position of all reference holes \(^{B}_{B}\vec{P}_{h^i}, i=[1,..,N_i]\), the orientation of the workpiece is found by fitting the estimated holes positions against a pre-known model of the workpiece; which can be done using an Iterative Closest Point approach. In our case, we only assume that the workpiece is flat such that all reference holes lie on the same plane; and hence we do not require knowledge of the number or position of holes in the workpiece. We simply fit a plane through the estimated position of all reference holes and and infer the workpiece's orientation from the parameters of the fitted plane. This plane-fitting step is also used to remove any outliers or noise in the localization of holes.

\section{Neuromorphic Event-based Hole Detection and Tracking}
\label{sec_hole_det}

The precise detection of circular holes from visual feedback directly affects the positional accuracy of the drilling process \cite{Xia2020}. In frame-based vision, several Well established methods exist for detecting circular formations in images \cite{Xia2020, Mei2015, Lou2020}, and the Circle Hough Transform (CHT) is amongst the most popular of these methods \cite{Atherton1999, Yuen1990}. In this section, we present an event-based variant of CHT that is appropriate for the asynchronous output of neuromorphic cameras

\begin{figure}[!t]
 \centering
\includegraphics[width=\linewidth]{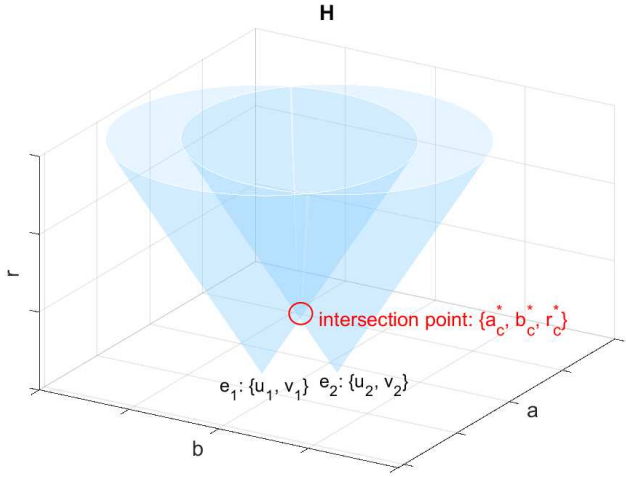}
\caption{The circle hough parameter space \(H\). Each edge point \(e_i\) in the image plane represents a hollow cone in the hough parameter space. The intersection of multiple cones signals the presence of a circular formation in the image plane.}
\label{fig_houh_detection}
\vspace{-1em}
\end{figure}

In conventional CHT, each feature point (e.g. edge point) in the image frame is mapped to the hough parameter space using the constraint equation given below:
\begin{equation}
    (u_i - a)^2 + (v_i - b)^2 = r^2 
    \label{eq_cht}
\end{equation}
where \((u_i, v_i)\) are the pixel coordinates of the i'th edge point, \((a, b)\) are the coordinates of circle's center, and \(r\) is the circle's radius. As such, a three dimensional parameter space \(H \in \mathbb{R}^3\), often referred to as the hough parameter space, that spans all possible values for \(a\), \(b\) and \(r\) is defined. Following \eqref{eq_cht}, each edge point in the image plane represents a hollow cone in \(H\) as depicted in Figure \ref{fig_houh_detection}. The intersection of multiple cones signals the presence of a circle with parameters that correspond to the intersection's location in \(H\). In practice, \(H\) is discritized to form an accumulator array \(\bar{H}(a, b, r)\), and each edge point contributes to \(\bar{H}\) through a voting process. Circle parameters are finally extracted from peaks in \(\bar{H}\).

CHT cannot be directly used with neuromorphic cameras due to their significantly different output from frame-based cameras. CHT establishes correspondence between edge points assuming that they are extracted from the same image frame with exact temporal match. Neuromorphic cameras on the other hand do not output image frames, but output an asynchronous stream of events in continuous time as shown in Figure \ref{fig_events_Vs_frames}. A naive solution would be to concatenate events within a defined time period to form artificial frames, and apply CHT to these frames. However, the generation of events is dependent on the rate of changes in the visual scene. As such, it would be challenging to determine a single period for event concatenation that is appropriate for all conditions. Figure. \ref{fig_circle_detector_different_detector} provides a visualization for this premise, where grouping events at different rates result in contradicting CHT performance at different egomotion velocities. 

To address these challenges, we formulate an event-based variant of CHT that adopts a bayesian framework to retain the asynchronous nature of neuromorphic cameras. In our algorithm, the accumulator array \(\bar{H}(a,b,r)\) is considered to be a Probability Mass Function (PMF) that reflects the probability of the existence of a circle for any given values of \(a\), \(b\), or \(r\). Following the principle of recursive bayesian filtering, our approach for event-based circle detection follows two steps: measurement update and prediction. The measurement update step follows the traditional CHT but in an asynchronous manner, where each event independently votes to regions in \(\bar{H}\) that satisfy \eqref{eq_cht}. Measurement update is performed whenever a new event is received, and is normalized so that \(\sum\bar{H} = 1\) after each update step.

The prediction step establishes a temporal continuity between events triggered at different times, and allows for the inference of circle parameters at any point in continuous time. In the prediction step, we update \(\bar{H}\) using the camera's egomotion, which is obtained by solving the forward kinematics of the robot manipulator as described in section \ref{sec_setup}. Given the camera's twist vector \(\vec{\mathcal{V}}_c = [v_x, v_y, v_z, \omega_x, \omega_y, \omega_z]^T\) that describe the camera's velocity in \(\mathcal{F}_C\), where \((v_x, v_y, v_z)\) are the linear components and \((\omega_x, \omega_y, \omega_z)\) are the angular components; the velocity in pixel coordinates of a feature point \((u, v)\) can be computed using the image Jacobian as shown below:
\begin{equation}
    \begin{bmatrix} \dot{u} \\ \dot{v} \end{bmatrix} = \begin{bmatrix}
    \frac{-F}{Z} & 0 & u/Z & uv/F & -(F+u^2/F) & v \\
    0 & \frac{-F}{Z} & v/Z & F+v^2/F & -uv/F & -u
    \end{bmatrix} \vec{V}_c 
    \label{eq_img_jac}
\end{equation}
Where \(F\) is the focal length, and \(Z\) is depth. In our control pipeline discussed in section \ref{sec_control}, we constrain the camera' motion during the reference hole detection step to a linear 2D motion perpendicular to the camera's optic axis, such that \(v_z = 0\) and \(\omega_x = \omega_y = \omega_z = 0\). This constraint simplifies the expression in \eqref{eq_img_jac} to:
\begin{equation}
    \begin{bmatrix} \dot{u} \\ \dot{v} \end{bmatrix} = \begin{bmatrix}
    \frac{-F}{Z} & 0 \\
    0 & \frac{-F}{Z} 
    \end{bmatrix} 
    \begin{bmatrix}
    v_x \\ v_y
    \end{bmatrix}
    \label{eq_img_jac_2D}
\end{equation}

It is evident from \eqref{eq_img_jac_2D} that features' velocities in pixel coordinates are uniform across all pixel locations, since they only depend on the camera's velocity and the depth of the point. We denote this uniform velocity by \((\dot{u}_{avg},\dot{v}_{avg})\). This translates to uniform motion in the \(a\) and \(b\) components of \(\bar{H}\). We define a gaussian kernel \(g(u, v, r)\) that incorporates this motion in accumulator array while also as follows: 
\begin{equation}
    g(u,v,r) = \alpha \hspace{1pt} exp\begin{pmatrix} \frac{1}{2}
    \mathbf{\bar{X}}
    \mathbf{\Sigma}^{-1}
    \mathbf{\bar{X}}
    \end{pmatrix}
    \label{eq_gaussian_filter}
\end{equation}
\begin{equation}
    \mathbf{\bar{X}} = \begin{pmatrix}
    \begin{bmatrix}
    u \\ v \\ r
    \end{bmatrix}
    - 
    \begin{bmatrix}
    \Delta t \dot{u}_{avg} \\ \Delta t \dot{v}_{avg} \\ 0
    \end{bmatrix}
    \end{pmatrix}
    \label{eq_gaussian_filter_mean}
\end{equation}
Where \(\Delta t\) is the time since the last prediction step, \(\mathbf{\Sigma}\) is the tunable covariance matrix, and \(\alpha\) is a scale factor so that \(\sum g = 1\). Although the expression in \eqref{eq_img_jac_2D} is deterministic, we select a gaussian distribution to model uncertainties in \(F\), \(Z\), or the camera's egomotion. 

The prediction step is then realized by the convolution of \(\bar{H}\) with \(g\) as shown in \eqref{eq_prediction_step}. This step can be performed whenever an event is triggered or at a fixed rate independent from event generation. In our experiments, the prediction step is applied at a rate of 100Hz. 
\begin{equation}
    \bar{H}(t^+) = \bar{H}(t^-) * g
    \label{eq_prediction_step}
\end{equation}

Finally, the circle parameters are extracted from the highest probability region of \(\bar{H}\) as follows:
\begin{equation}
    \begin{bmatrix}
    a_{c}^{*} & b_{c}^{*} & r_{c}^{*}
    \end{bmatrix}
    = 
    \argmax_{(a,b,r)} \bar{H}
    \label{eq_inference_step}
\end{equation}

\begin{figure}[!t]
 \centering
\fbox{\includegraphics[width=0.8\linewidth]{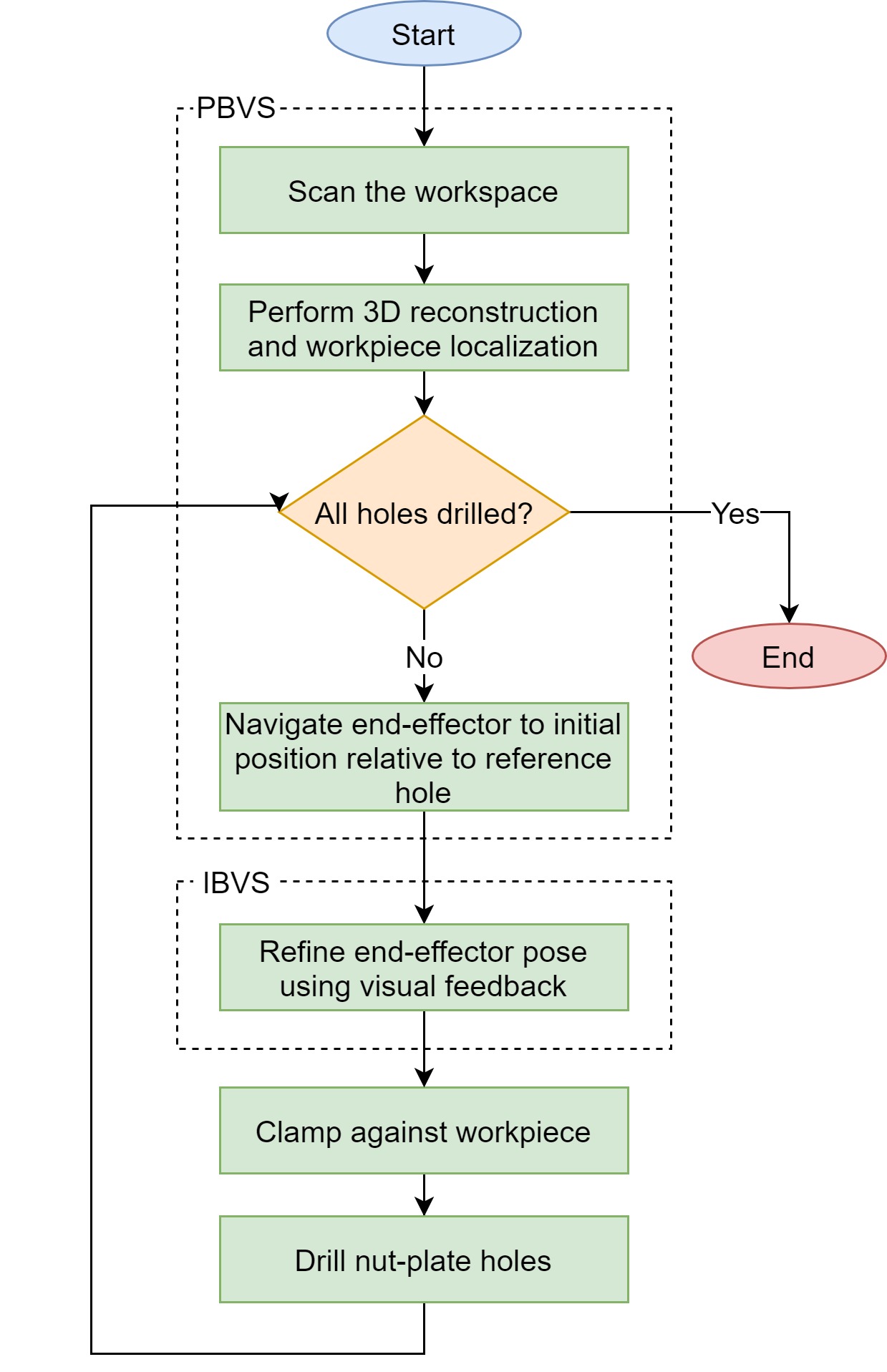}}
\caption{Outline of the different control steps for the proposed neuromorphic vision-based drilling system. The robot performs multiple stages of PBVS and IBVS to scan the environment, align with the wokrpiece, and drill the desired holes. This process is repeated until all the holes on the workpiece are drilled.}
\label{fig_robot_control}
\vspace{-1em}
\end{figure}

\section{Vision Based Robot Controller}
\label{sec_control}

This section explains the vision-based control logic that utilizes the perception algorithms in sections \ref{sec_multi_view} and \ref{sec_hole_det} to regulate the robot motion during the drilling procedure. Figure \ref{fig_robot_control} provides an outline of this controller, which consists of two subsequent PBVS and IBVS stages. The PBVS stage guides the end-effector towards initial alignment with the reference holes on the workpiece using the 6-DoF pose estimate from the multi-view detection. IBVS refines the end-effector alignment to sub-millimeter accuracy using the event-based hole detection algorithm. Both stages are described in detail in sections \ref{sec_PBVS} and \ref{sec_IBVS}.

\subsection{Position Based Robot Control (PBVS)}
\label{sec_PBVS}
In the PBVS stage, we consider a known desired pose for the robot's end effector denoted by \(_{B}\hat{T}_{E}\). This pose is either a pre-defined constant, which is the case during the scanning step; or is computed from knowledge of the reference holes' poses \(_{B}T_{h_i}\) and a pre-defined stance of the end-effector relative to these holes \(_{h_{i}}\hat{T}_{E}\). Concurrently, we define a desired joint angles vector \(\hat{\theta}\in\mathbb{C}\) such that: 
\begin{equation}
    \label{eq_forward_kinematric_desired}
    _{B}\hat{T}_{E} = g(\hat{\theta})
\end{equation}
which we solve for using the Newton-Raphson inverse kinematic approach of the open-source Kinematic and Dynamics Library (KDL)\footnote{KDL: https://www.orocos.org/kdl.html}. Using the current joint angles \(\theta\) and the desired ones \(\hat{\theta}\), we compute a time-parametrized trajectory for the joint angles \(\hat{\theta}(t)\) using RRT-connect \cite{rrt2000} implementation on the Open Motion Planning Library \cite{ompl2012} of MoveIt!\footnote{MoveIt!: https://moveit.ros.org/}. Finally, a low-level PID controller regulates each joint to track \(\hat{\theta}(t)\).

\subsection{Image Based Robot Control (IBVS)}
\label{sec_IBVS}

The IBVS stage refines the end-effector's position based on the detected hole location in image coordinates. Let ${\vec{f} = [a_{c}^{*}, b_{c}^{*}]^T} \in \mathbb{R}^2 $ denote the pixel coordinates of the the detected hole, and  $\vec{\hat{f}} \in \mathbb{R}^2$ denote the desired coordinates of these high level features (e.g. the camera's principal point); we define an error vector as $\vec{\zeta}=  \vec{f} - \vec{\hat{f}}$, and a control law that exponentially decays this error to zero as:
\begin{equation}
 \dot{\vec{\zeta}} = - \lambda \vec{\zeta}
 \label{eq_ibvs_control_law}
\end{equation}
where $\lambda \in \mathbb{R}^{2 \times 2}$ is a positive-definite gain matrix.


What follows is the generation of joint movements that achieve the desired $\dot\vec{\zeta}$. First, we define the command twist vector \(\vec{V}^*_c = [v^*_x, v^*_y, v^*_z, \omega^*_x, \omega^*_y, \omega^*_z]^T\) that describes the camera's desired velocity in \(\mathcal{F}_C\). In our case, We constrain the camera's motion to a linear 2D movement perpendicular to camera's optical axis, such that \([v^*_z = 0]\) and \([\omega^*_x, \omega^*_y, \omega^*_z] = \vec{0}\). Hence, the linear \(x\) and \(y\) components of \(V^*_c\) can be easily computed by inverting the expression in \eqref{eq_img_jac_2D}, as follows:
\begin{equation}
    \label{eq_ibvs_control_v}
    \begin{bmatrix} 
    v^*_x \\ v^*_y
    \end{bmatrix} = \begin{bmatrix}
    -Z/F & 0 \\
    0 & -Z/F
    \end{bmatrix} \dot{\vec{\zeta}}
\end{equation}
It must be noted that the depth value \(Z\) is obtained from \(_{C}T_{h_i} = _{B}T_{C}^{-1} \hspace{0.5em} _{B}T_{h^i}^{-1}\), where \(_{B}T_{h^i}\) is estimated from the multi-view 3D localization step.

Reference joint angular velocities \(\dot{\theta}^*\) that result in \(\vec{V}^*_c\) are then calculated by inverting the expressions in \eqref{eq_vel_forward_kinematics} and \eqref{eq_adjoint_representation} as:
\begin{equation}
    \dot{\theta}^* = J(\theta)^{\dagger} \hspace{0.5em} _{C}[AD_T]_{E}^{-1} \hspace{0.5em} \vec{V}^*_c
    \label{eq_ibvs_inverse_kinematics}
\end{equation}
 Finally, \(\dot{\theta}^*\) is tracked using PID control for each indivdual joint.

\begin{figure*}[T]
      \centering
      \subfloat[]{\label{fig:sf1}
      \includegraphics[height=0.4\textwidth]{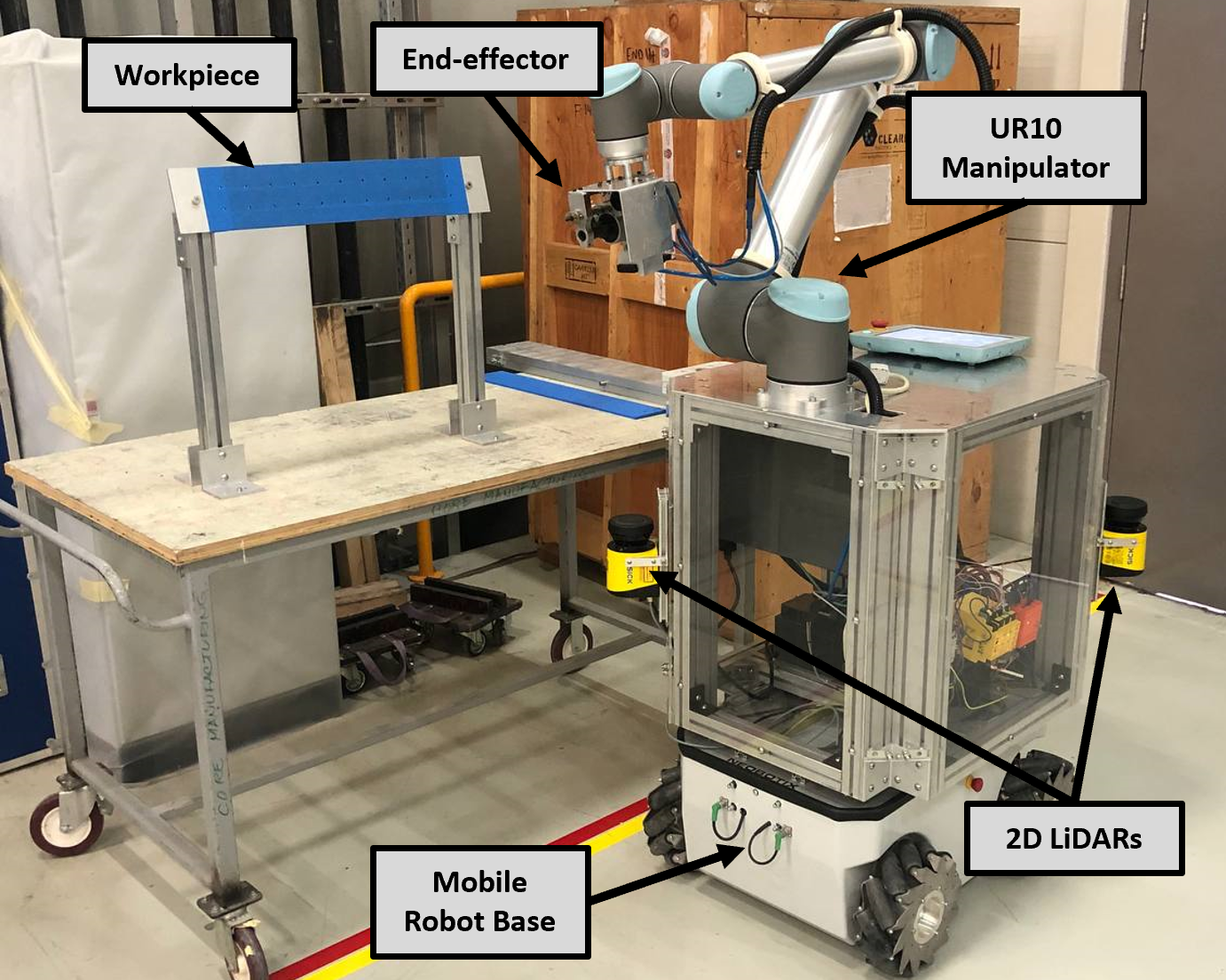}}  \qquad
      \subfloat[]{\label{fig:sf1}
      \includegraphics[height=0.33\textwidth]{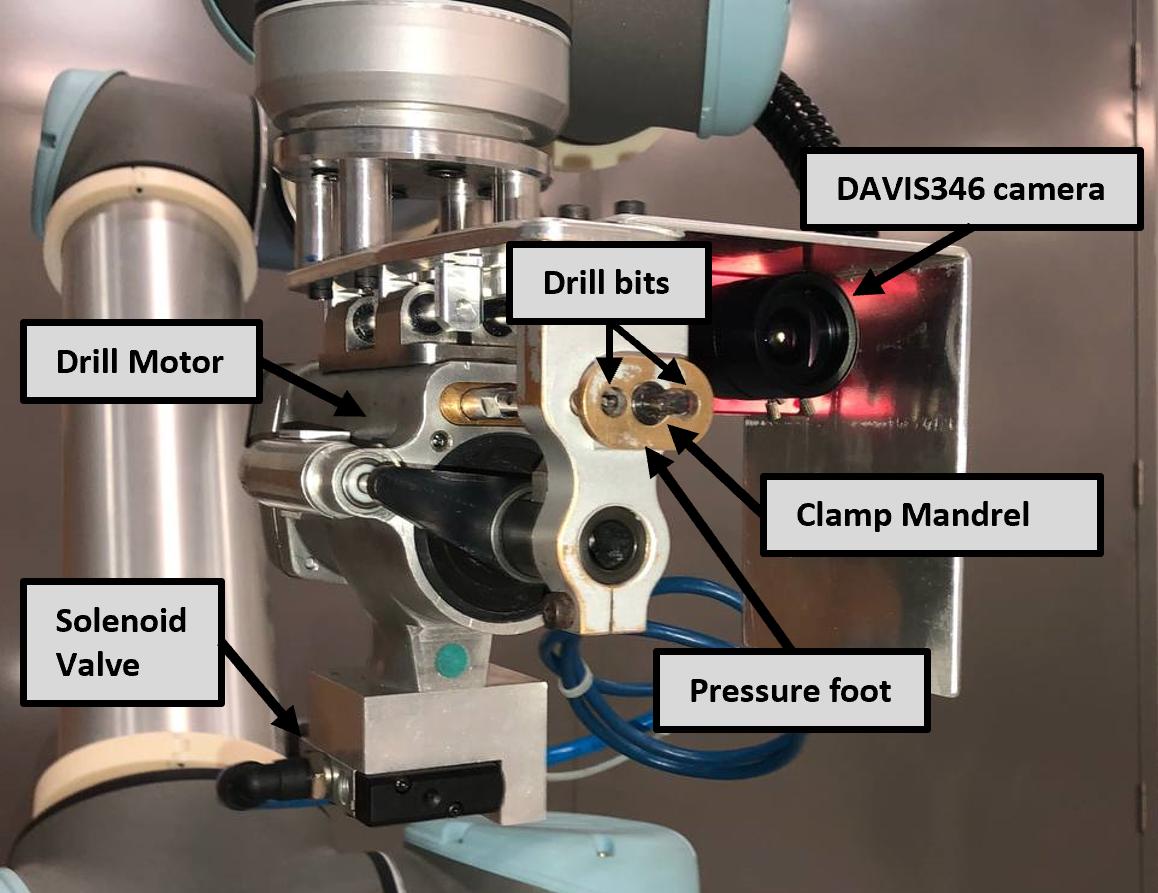}}  \qquad
    \caption{Experimental setup for testing the proposed neuromorphic vision-based drilling method. (a) The mobile robot and the manipulator setup. (b) The configuration of the end-effector comprising the visual sensor and the drill motor.}
    \label{fig_robot_exp_setup}
\end{figure*}

Once the robot's end-effector is aligned with the target reference hole on the workpiece, the clamp mandrel (see Fig. \ref{fig_robot_exp_setup}-b) is inserted in the reference hole and the pressure foot clamps up against the workpiece until a target contact force is achieved. These contact forces are estimated from the torques on each of the robot's joints. Afterwards, the clamp mandrel is retracted against the workpiece from the blind side, providing additional clamping force. This two-sided clamping ensures stability during the drilling process and minimizes normality errors using the inherent compliance of the robot manipulator. The robot then proceeds with activating the drill motor and drilling nutplate installation holes on the sides of the reference hole.

\section{Experimental Validation and Results}
\label{sec_exp}

\subsection{Experimental Setup and Protocol}
The presented event-based robotic drilling algorithms were tested on the setup shown in Figure \ref{fig_robot_exp_setup}. We used Universal Robot's UR10 \cite{UR10}, which provides a repeatability of 0.1mm, as the primary manipulator. The manipulator was mounted on top of a customized version of the Neobotix MPO-500 robot base \cite{mpo500}. The mobile robot uses two Sick S300 LIDARs \cite{sick300} and the ROS Navigation Stack \cite{ros_navi} to  autonomously navigate the factory settings, and place the manipulator in the vicinity of the workpiece. This enables the robot to perform multiple drilling jobs and operate across a large workspace without the need for human involvement. Figure \ref{fig_robot_exp_setup}-b displays the end-effector configuration, which comprises the drill-motor and the camera. Inivation's DAVIS346 camera \cite{davis346} with a spatial resolution of 346x260 was used for visual perception, which provides both a neuromorphic event stream in addition to conventional frame-based intensity images. The event stream grants a dynamic range of 120 dB, a latency of \( ~ 20 \mu s\), and a bandwidth of \(12\times10^{6}\) events per second. While all operations are conducted solely using the event stream, intensity images are used as a benchmark to assess the performance and advantages of event-based perception. All required computations are executed using an on-board computer with an i7-5530 processer and 4GB of RAM.

Our experimental evaluation focuses on three aspects. In section \ref{sec_exp_3D}, we assess the accuracy of the event-driven multi-view 3D reconstruction technique, and quantify its advantages against conventional frame-based vision under different conditions of lighting and operational speeds. Similarly, we evaluate and benchmark our event-based hole detection and tracking pipeline in section \ref{sec_exp_hole}. Finally, section \ref{sec_exp_drill} presents the evaluation data of the comprehensive drilling experiments.

\subsection{6-DOF Workpiece localization}
\label{sec_exp_3D}
\begin{figure*}[T]
      \centering
      \subfloat[]{\label{fig:sf1}
      \includegraphics[width=0.4\textwidth]{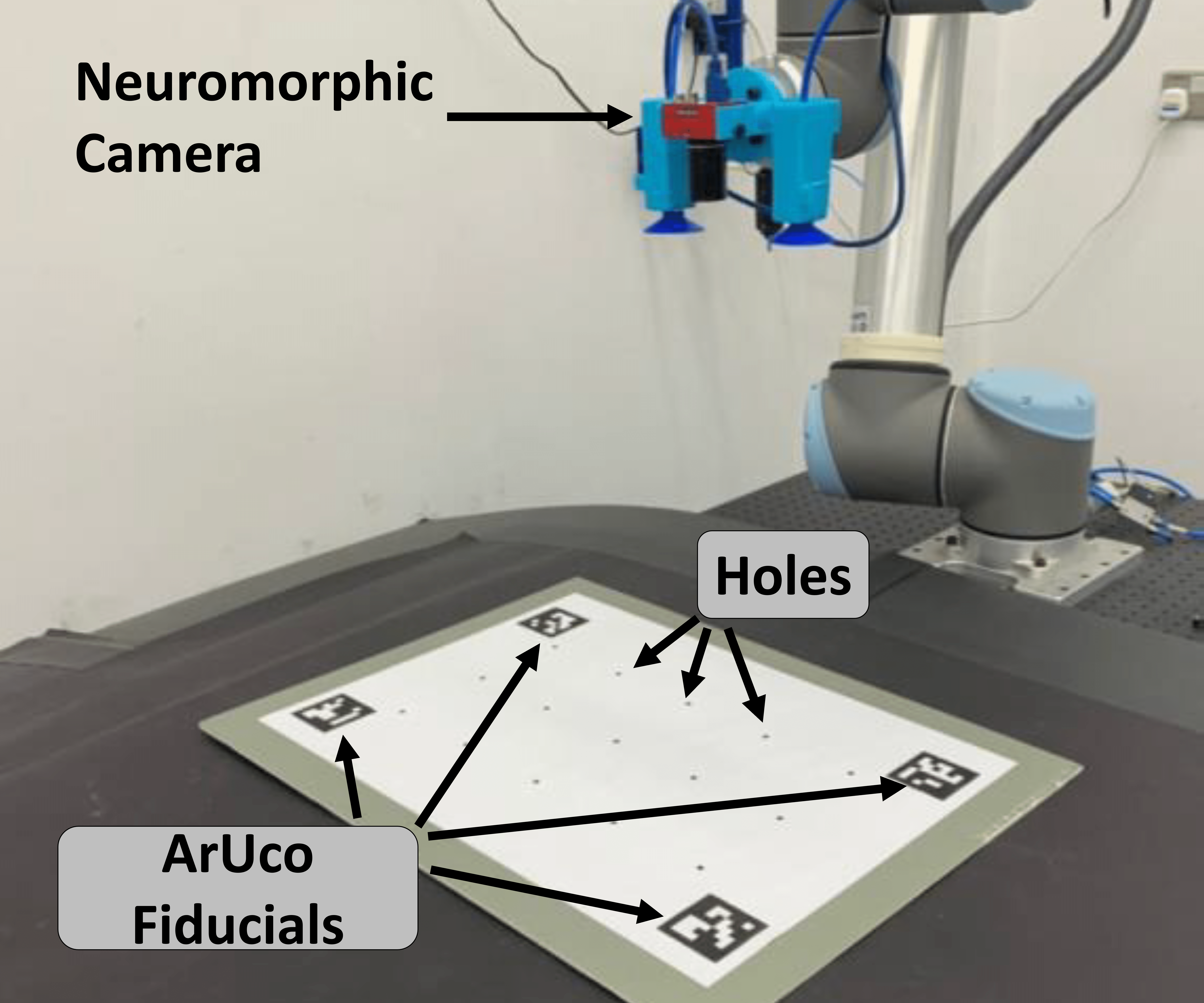}}  \qquad
      \subfloat[]{\label{fig:sf1}
      \includegraphics[width=0.4\textwidth]{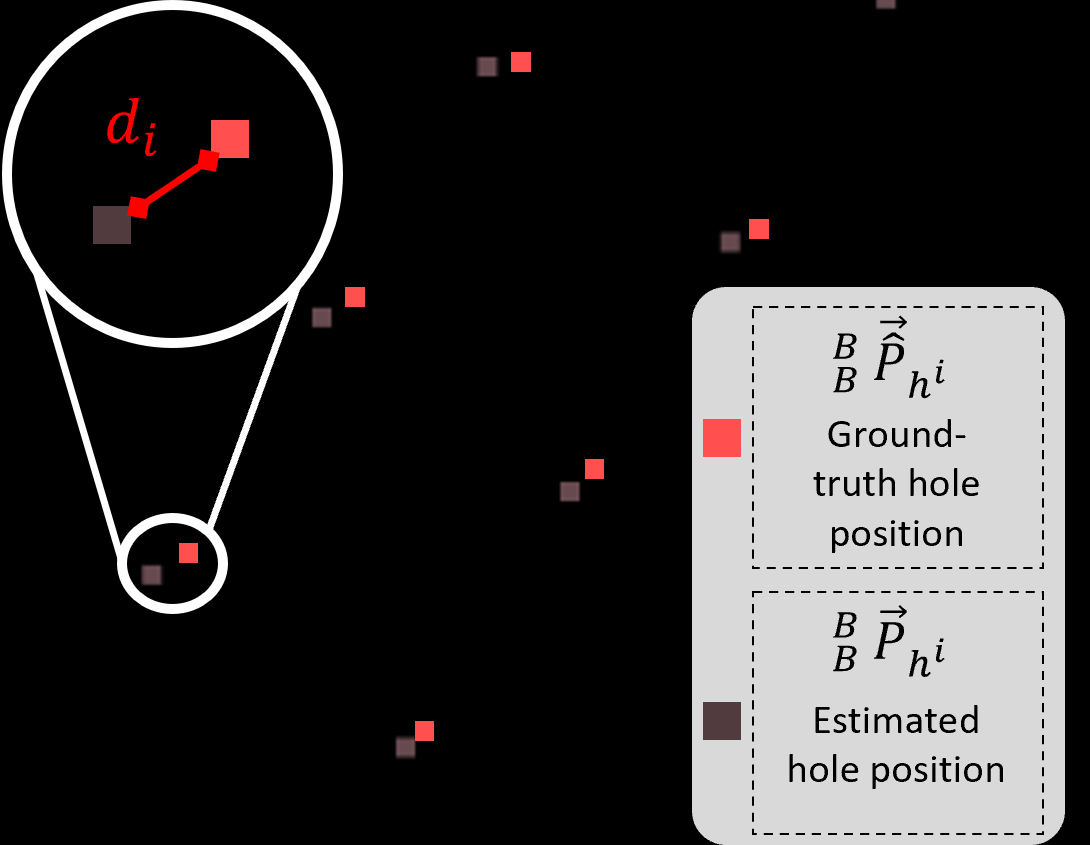}}  \qquad
    \caption{The experimental evaluation of the neuromorphic 3D workpiece localization pipeline. (a) The experimental setup. (b) The error is quantified in terms of the euclidean distance between the estimated hole location and the corresponding ground-truth location extracted using a set of AruCo Fiducials.}
    \label{fig_multi_view_eval_setup}
\end{figure*}

In this section, we quantify the accuracy of the proposed event-based multi-view reconstruction approach that provides initial estimates of position of the workpiece and its reference holes. The setup for these experiments is shown in Figure \ref{fig_multi_view_eval_setup}. Ground truth data is obtained using a set of four ArUco fiducials \cite{aruco_marker}, denoted by \(Ar_{j, j=1,...,4}\). The fiducials are observed from a static robot state using the intensity image output of the DAVIS346, and the 6-DoF pose of each fidicual \(_{B}T_{Ar^j}\) is estimated using OpenCV's ArUco library \cite{aruco_library}. As the position of each hole relative to each fiducial (denoted by \(^{Ar^j} _{Ar^j}\vec{P}_{h^i}\)) is known with high accuracy, the ground truth position of the i'th hole relative to \(\mathcal{F}_B\) is given by: 
\begin{equation}
    ^{B}_{B}\vec{\hat{P}}_{h^i} = \frac{1}{4}\sum_{j=1}^4 \hspace{1pt} {}_{B}T_{Ar^{j}} \hspace{1pt} ^{Ar^{j}} _{Ar^{j}}\vec{P}_{h^i}
    \label{eq_ground_truth_hole}
\end{equation}

Consequently, we define the error \(d_i\) in the 3D localization of each hole as the mismatch between the hole's estimated location \(^{B}_{B}\vec{P}_{h^i}\) obtained using the multi-view approach and its ground truth location \(^{B}_{B}\vec{\hat{P}}_{h^i}\). This mismatch is averaged across all holes in the workpiece to quantify the overall error in 3D workpiece localization as follows:

\begin{equation}
    D = \frac{1}{N_i} \sum_{i=1}^{N_i} d_i , \hspace{0.4em}
    \\ \\
    d_i = || ^{B}_{B}\vec{P}_{h^i} \hspace{0.3em} - ^{B}_{B}\vec{\hat{P}}_{h^i} ||
    \label{eq_workpiece_err}
\end{equation}
where \(N_i\) represent the overall number of reference holes.

We benchmark the multi-view localization results obtained using the neuromorphic event stream against results obtained using conventional intentsity images. For conventional images, we use the standard Canny edge detector to extract features from each image and then perform the same multi-view procedure as described in section \ref{sec_multi_view}. To better assess the advantages of neuromorphic vision, we conduct the multi-view localization experiments across different lighting conditions and various speeds of the workspace scanning movement. Table \ref{tab_emvs_results} summarizes the obtained results in terms of position error across all experimental condition for both the event-based and conventional image-based approaches.

\renewcommand{\arraystretch}{1.5}
\begin{table*}[H]
\caption{Evaluatuin of the 3D workpiece localization error $D$ in $mm$ using both neuromorphic and conventional frame-based vision under different light intensity levels $I_v$ and maximum robot speed during the scanning movement $v_{max}$. Bold indicates the lower error. These positional errors are then reduced by the IBVS stage to sub-millimeter errors as seen in Table. \ref{tab_drilling_pos_result}}
\centering
\begin{tabular}{|l||c|c||c|c|}
\hline
\backslashbox{\begin{tabular}{@{}c@{}}\textbf{\(v_{max}\)} \\  ($m/s$)\end{tabular}}{\begin{tabular}{@{}c@{}}\textbf{\(I_v\)} ($lx$)\end{tabular}} & 
\multicolumn{2}{c||}{$\sim 400$ (Adequate lighting)} & 
\multicolumn{2}{c|}{$< 10$ (low light)}  \\ 

& Neuromrophic  & Conventional & Neuromrophic & Conventional \\ \hline \hline

0.1    & \(2.97\)  & \(\mathbf{2.88}\)  & \(\mathbf{2.18}\)  & Fail \\ \hline
0.3     & \(3.19\)  & \(\mathbf{3.11}\)  & \(\mathbf{3.78}\) & Fail       \\  \hline
0.5      &  \(3.45\)  & \(\mathbf{3.27}\)  & \(\mathbf{7.36}\) & Fail       \\  \hline
1.0      &  \(\mathbf{2.77}\)  & \(3.71\)  & \(\mathbf{6.48}\) & Fail       \\  \hline
1.5      &  \(\mathbf{2.23}\)  & \(Fail\)  & \(\mathbf{8.73}\) & Fail       \\  \hline
\end{tabular}
\label{tab_emvs_results}
\end{table*}

\begin{figure}[!h]
 \centering
      \includegraphics[width=0.8\linewidth]{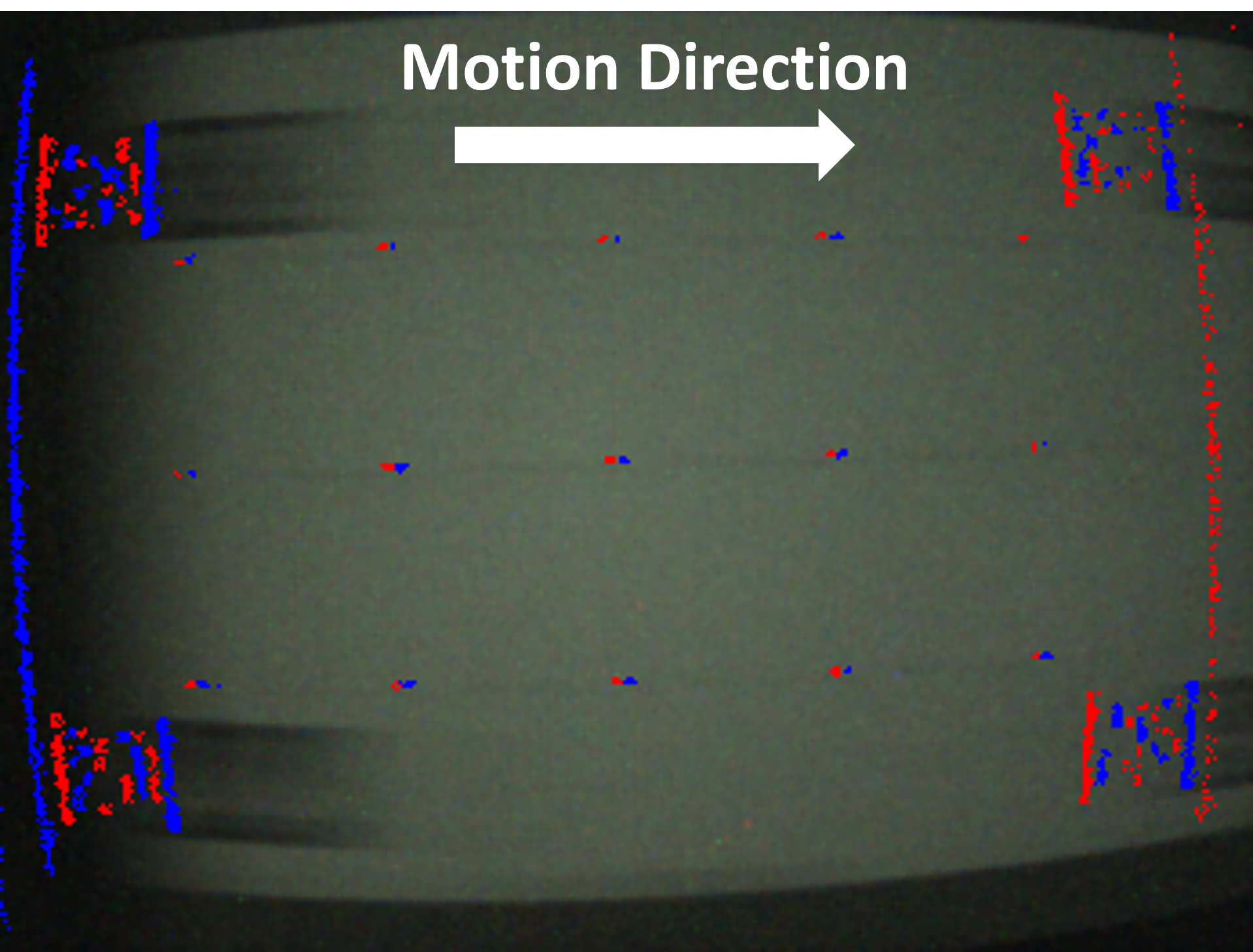}
\caption{The event stream (red and blue) and the intensity image output of DAVIS346 camera during the multi-view workpiece localization experiments on the setup shown in Figure \ref{fig_multi_view_eval_setup}. Intensity images clearly exhibit motion blur and higher latency while the event stream remains sharp and timely.}
\label{fig_multi_view_qualitative}
\vspace{-1em}
\end{figure}

Result in Table \ref{tab_emvs_results} indicate that at lower speeds and good lighting conditions, the multi-view localization using neuromorphic vision and conventional vision provides similar accuracy. The advantages of neuromorphic vision became apparent as the robot's operational speed is increased or at lower light conditions. At such conditions, perception using conventional cameras become challenging due to motion blur and high latency that results due to increased exposure timing. Neuromorphic cameras do not suffer from these shortcomings and as such, they result in more precise and reliable 3D localization results. Figure \ref{fig_multi_view_qualitative} visualizes the output of both types of cameras during the multi-view localization experiments, which confirms the reasoning of the superior performance of neuromorphic vision.

Regardless of the vision sensor, the multi-view localization step is not capable of achieving the sub-millimeter accuracy requirements of the drilling process. This signals the need for the second pose refinement step, which we do using hole detection, which we validate in the below section.

\begin{figure*}[B]
\centering
\begin{tabular}{m{0.1\textwidth}|>{\centering\arraybackslash}m{0.2\textwidth}|>{\centering\arraybackslash}m{0.2\textwidth}|>{\centering\arraybackslash}m{0.2\textwidth}|>{\centering\arraybackslash}m{0.2\textwidth}}

 & (a) Standard CHT using conventional images & (b) Standard CHT using event frames, \(\Delta t = 1 ms\) & (c) Standard CHT using event frames, \(\Delta t = 10 ms\) & (d) The proposed event-based CHT \\ \hline
 
 25 mm/s Egomotion & 
 \includegraphics[width=0.2\textwidth]{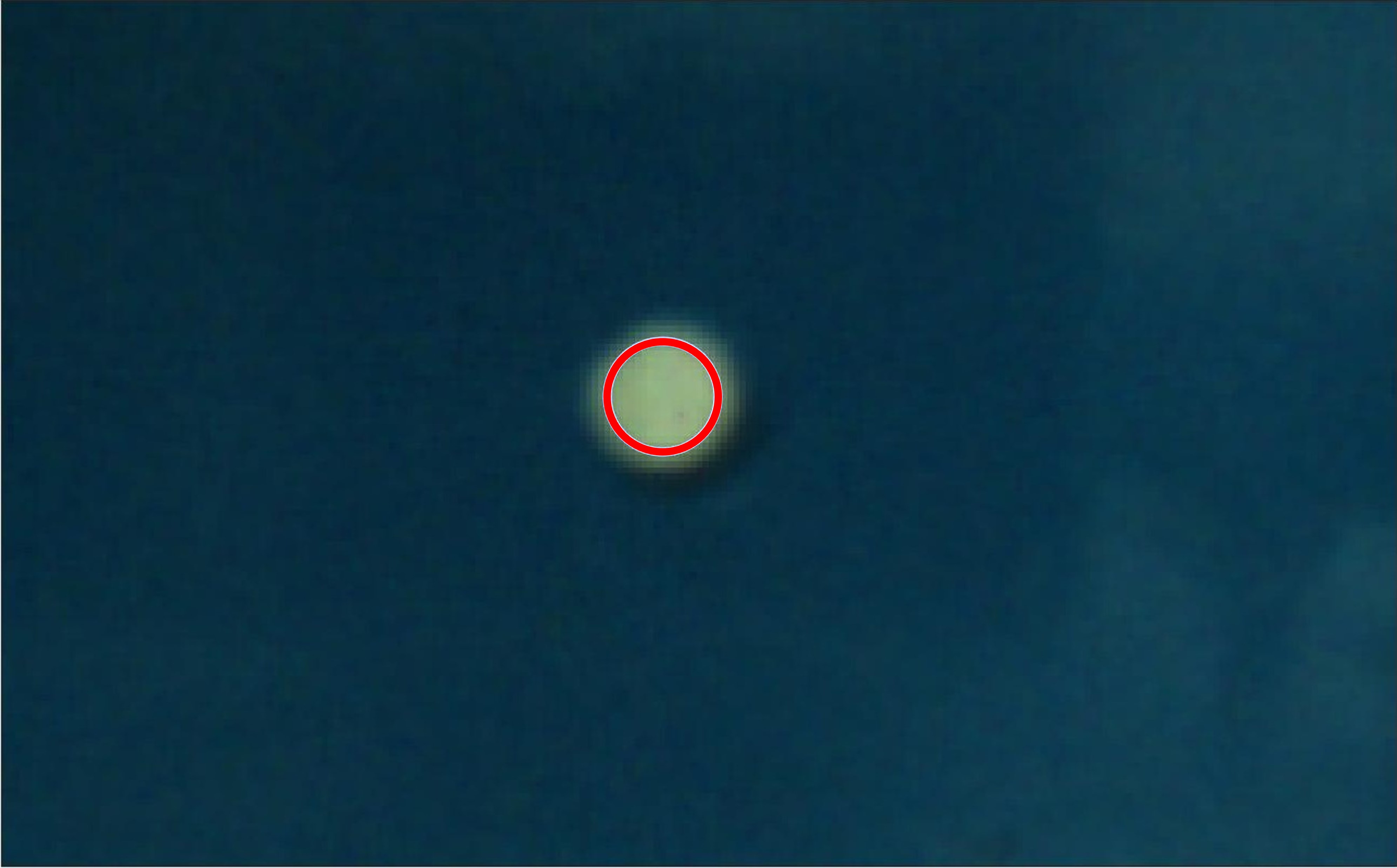} & 
 \includegraphics[width=0.2\textwidth]{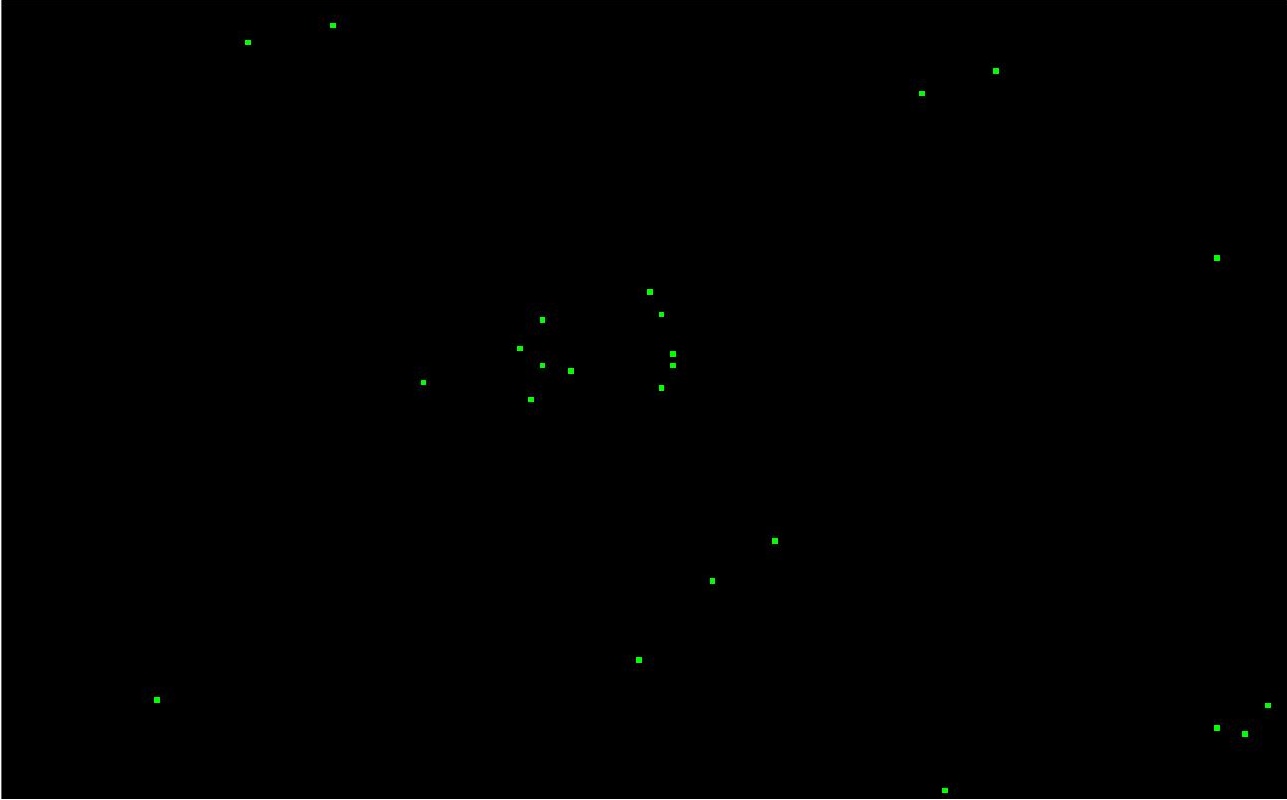} &
 \includegraphics[width=0.2\textwidth]{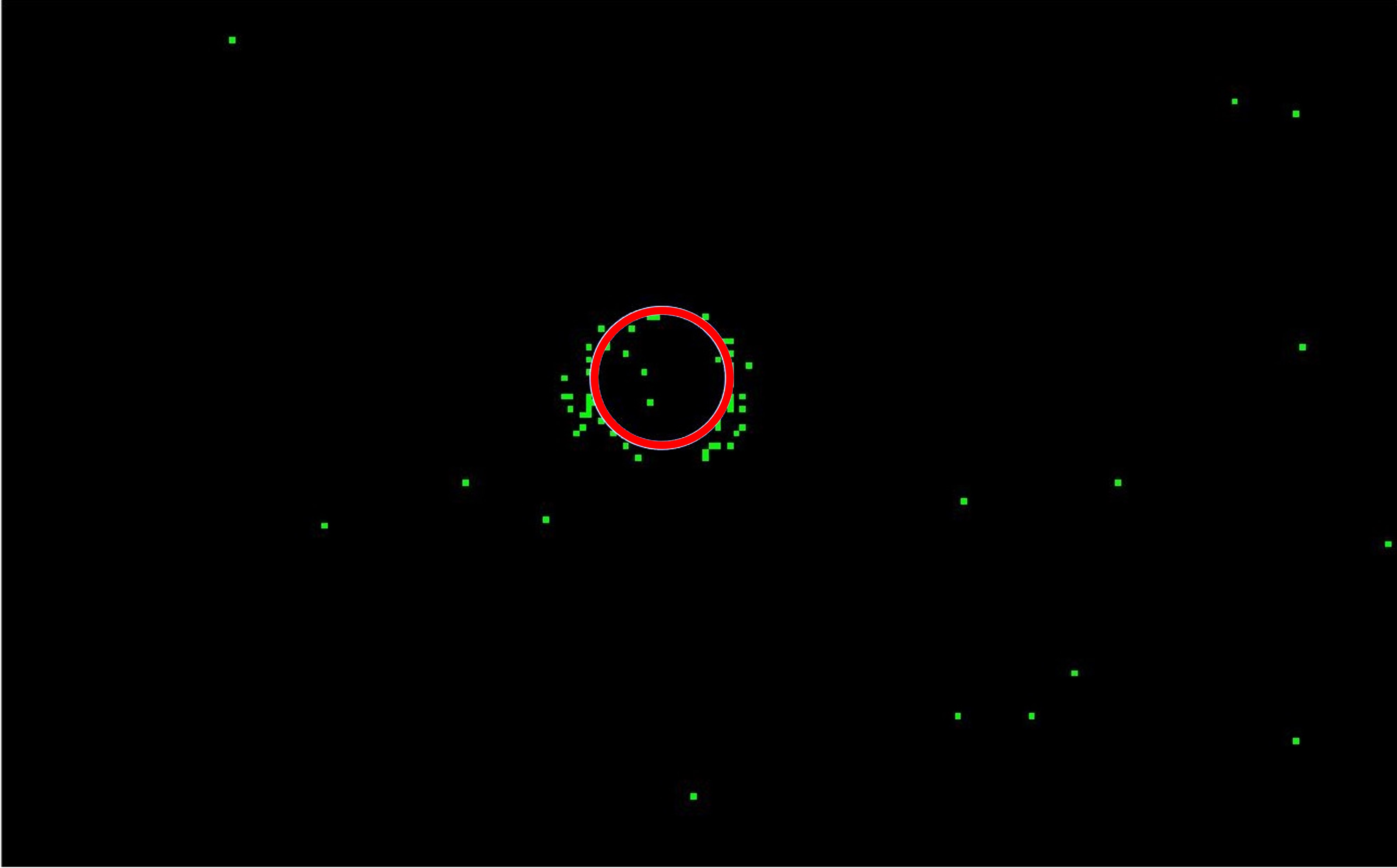} &
 \includegraphics[width=0.2\textwidth]{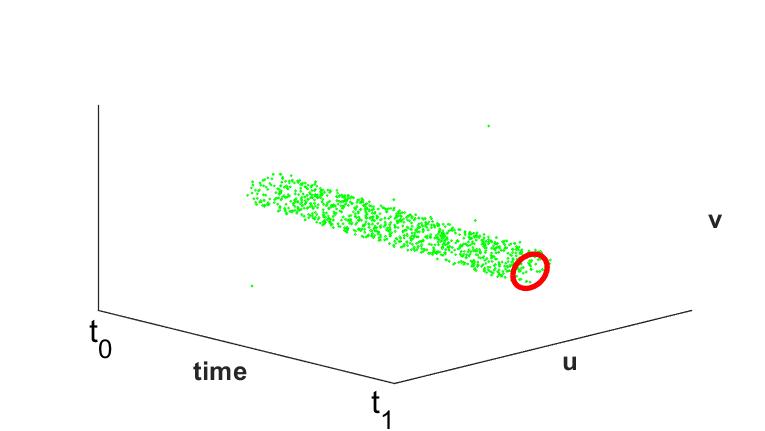} 
 
 \\ \hline
 200 mm/s Egomotion &
 \includegraphics[width=0.2\textwidth]{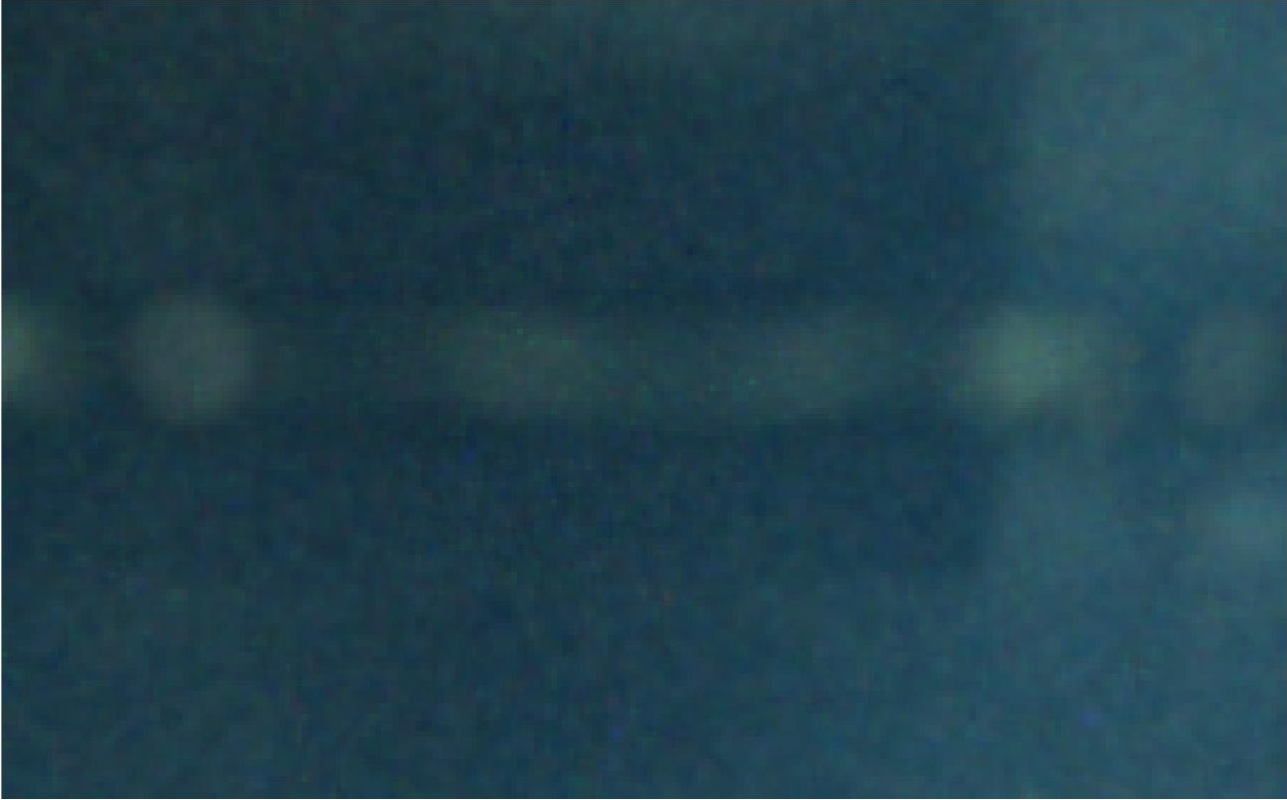} & 
 \includegraphics[width=0.2\textwidth]{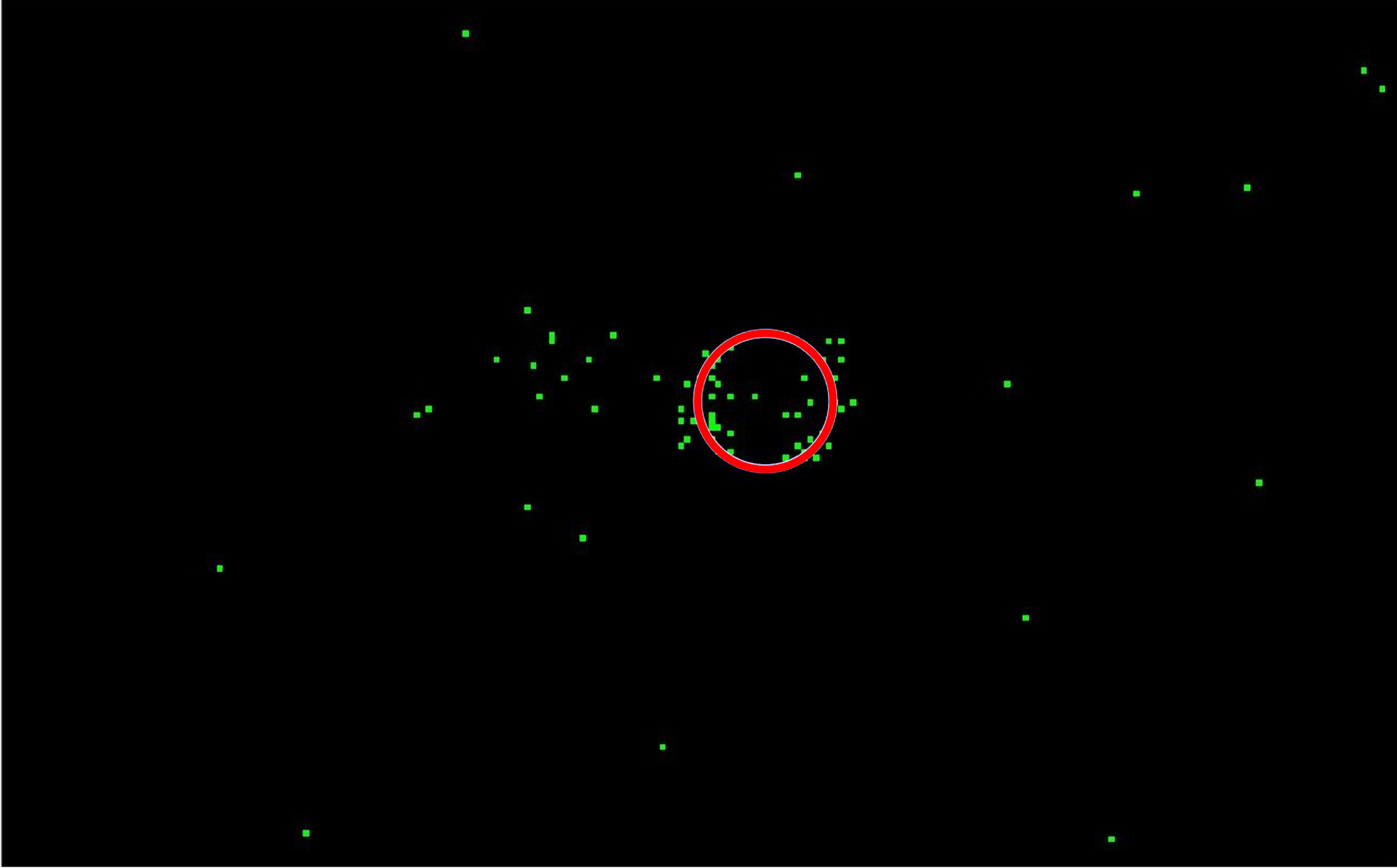} &
 \includegraphics[width=0.2\textwidth]{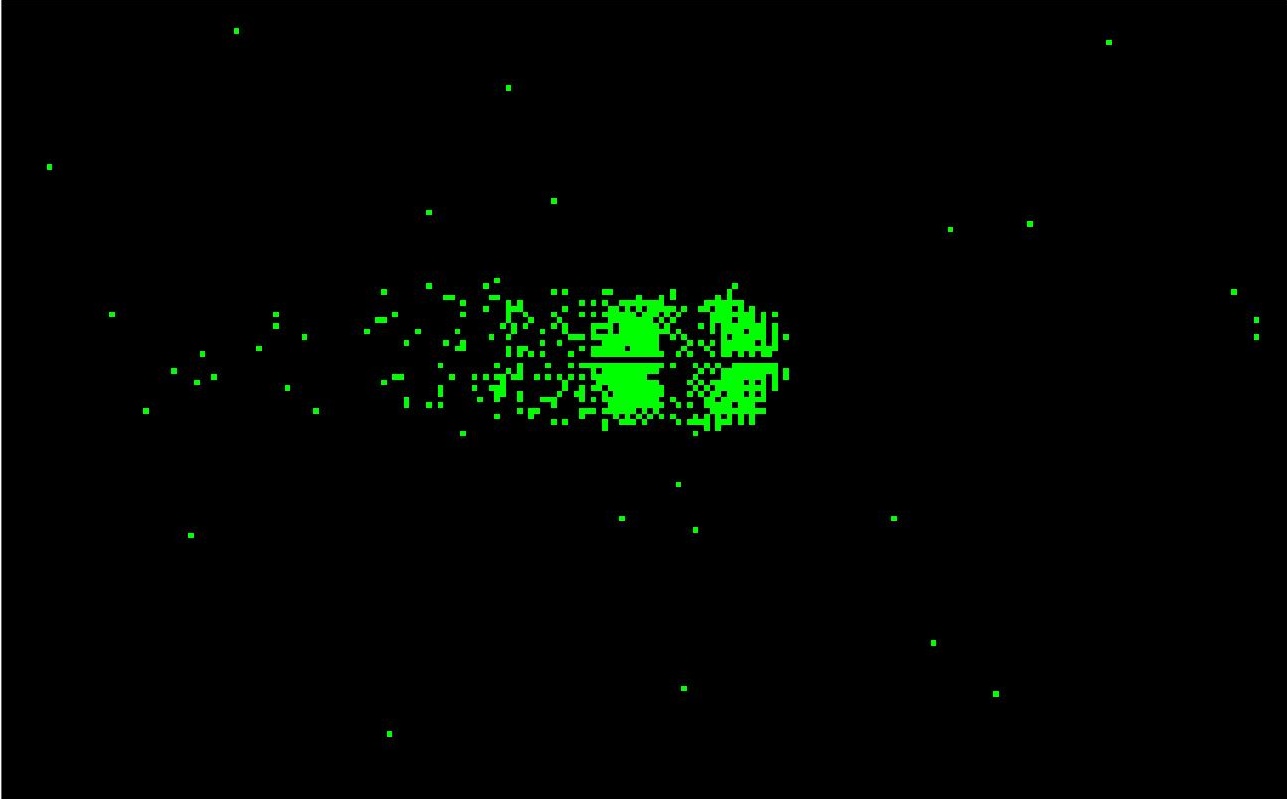} &
 \includegraphics[width=0.2\textwidth]{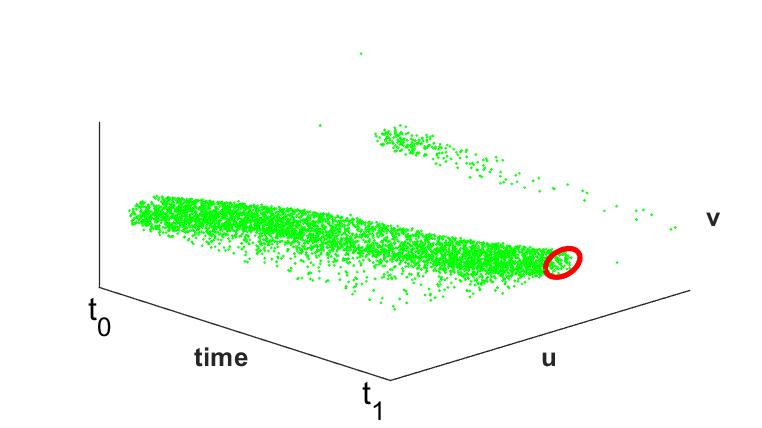}
 \\
 
 \multicolumn{5}{c}{\includegraphics[width=0.25\textwidth]{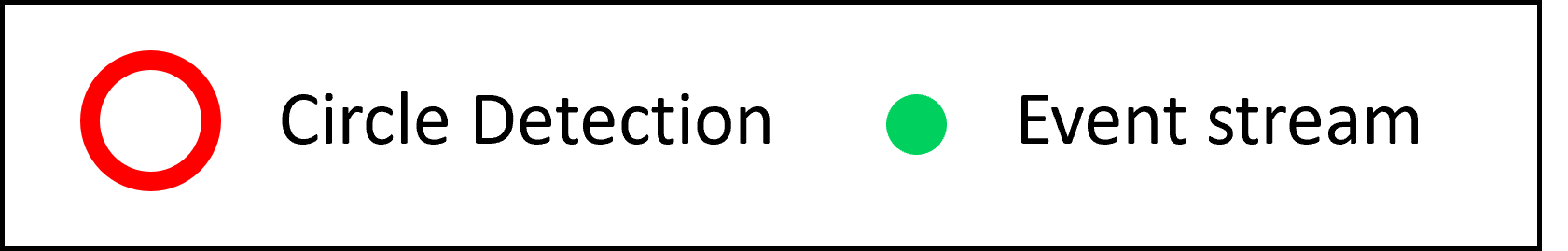}} 
 
 \\ 
\end{tabular}

\caption{Qualitative comparison between the proposed event-based variant of the CHT and the standard CHT applied to: (a) conventional images. (b) Artificial images created by the concatenation of events with a time period \(\Delta t = 1 ms\). (c) Artificial images created by the concatenation of events with a time period \(\Delta t = 1 ms\). The proposed event-based CHT provides consistent performance regardless of the camera's egomotion, while the performance of the standard CHT clearly degrades. In the case of regular images, the standard CHT fails due to excessive motion blur. As for event frames, the proper period at which events should be grouped is highly dependent on the motion in the visual scene, which in turn affects the quality of the CHT detection.}

\label{fig_circle_detector_different_detector}
\end{figure*}

\subsection{Neuromorphic Hole Detection}
\label{sec_exp_hole}

This section evaluates our proposed event-based circular detection and tracking method against the conventional CHT. For evaluation purposes, we test the conventional CHT with both intensity images and artificial image frames generated from the concatenation of events at specified time period \(\Delta t\). We assess each method's capability to track the circular holes in the workpiece under different movement speeds and light intensity levels. Figure \ref{fig_circle_detector_different_detector} visualizes the results of each method at different camera egomotions. The use of intensity images leads to unreliable detection at higher egomotion speeds due to motion blur in the visual scene, which is caused by the working principal of conventional imagers that rely on the time integration of incident light. Using conventional CHT with event frames is sensitive to the time period at which events are being grouped. For instance, a small concatenation period leads to a featureless image at slower speeds; while larger periods cause an over-populated image at high speeds, where feature can no longer be accurately extracted. Our event-based variant of the CHT exploits the advantages of neuromorphic vision and takes into the account the asynchronous nature of the event stream. As such, it provides precise results regardless of the motion in the scene and does not suffer from the motion blur or latency complications of conventional cameras.

\begin{figure*}[T]
\centering
\begin{tabular}{m{0.12\textwidth}|>{\centering\arraybackslash}m{0.425\textwidth}|>{\centering\arraybackslash}m{0.425\textwidth}}

 & Adequate light intensity ($\sim 400 lx$) &  Low light intensity ($< 1 lx$) \\ \hline
 
 Circle Detection and Tracking Results &
 \includegraphics[width=0.425\textwidth]{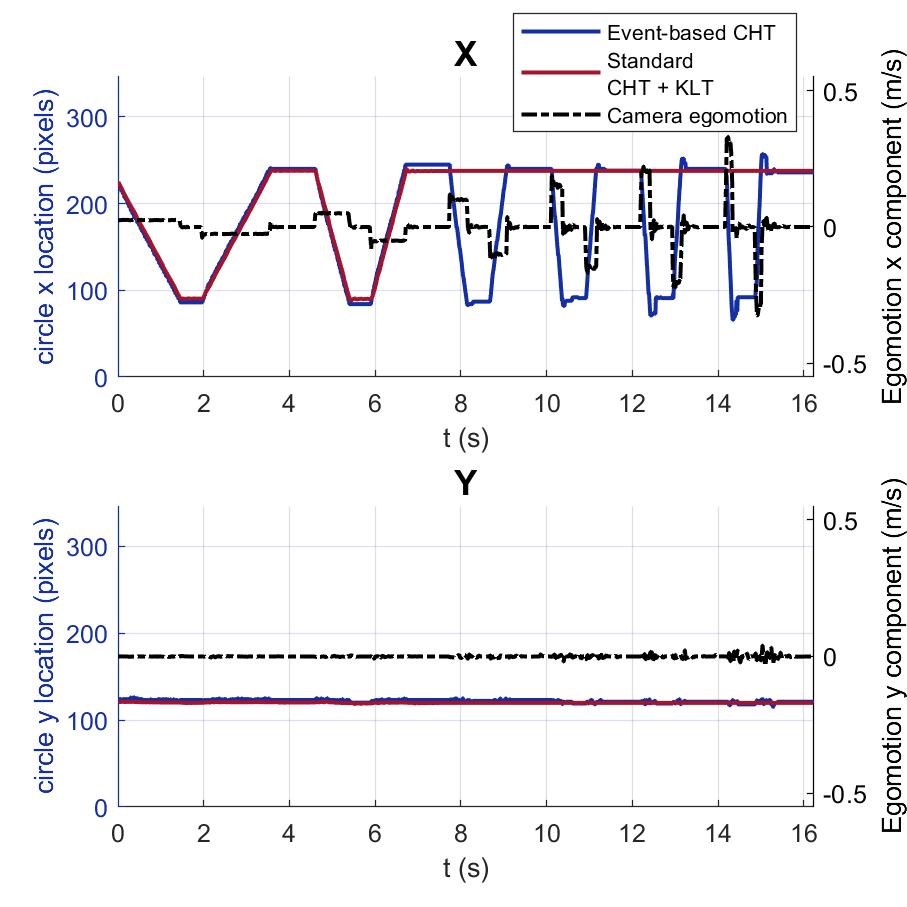}  &
 \includegraphics[width=0.425\textwidth]{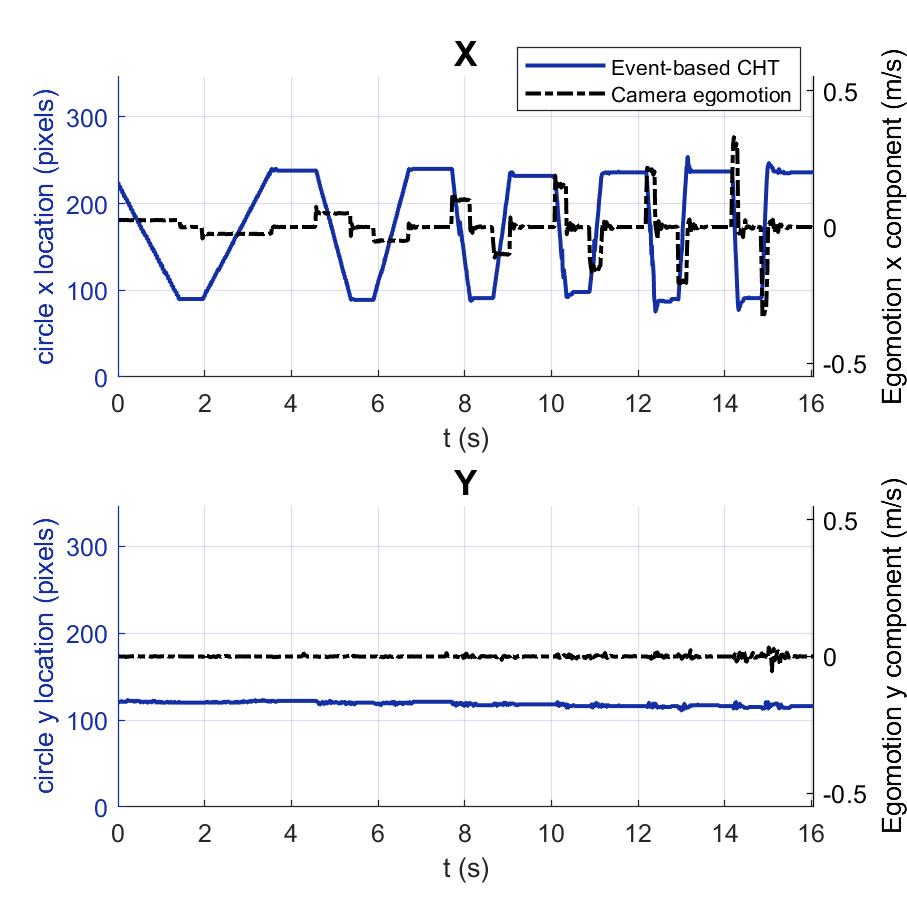}
 \\ \hline
 
 Camera View (Stationary camera)  &
 \includegraphics[width=0.3\textwidth]{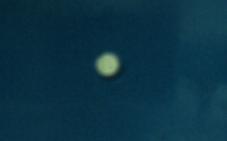}  &
 \includegraphics[width=0.3\textwidth]{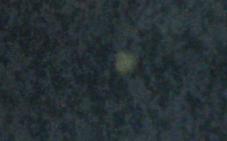}
 \\ \hline
 
 Camera View (Moving camera)  &
 \includegraphics[width=0.3\textwidth]{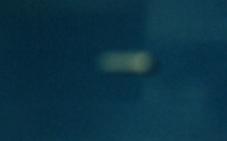}  &
 \includegraphics[width=0.3\textwidth]{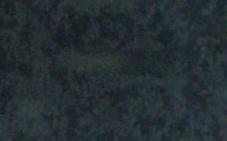}
 \\
 
 \\ 
\end{tabular}

\caption{Comparison between the proposed event-based variant of the CHT detector for the neuromorphic camera and the to conventional intensity image-based CHT with a KLT tracker for the detection and tracking of a circular hole 75mm in front of the camera at different lighting speeds and ego-motion velocities. The use of neuromorphic vision and our proposed circular hole detector clearly provides more reliable and consistent results at high operation speeds or imperfect lighting.}

\label{fig_circle_detector_different_speed}
\end{figure*}

The experimental results shown in figure \ref{fig_circle_detector_different_speed} further enforce the advantages of neuromorphic vision coupled with our proposed event-based CHT. In these experiments, the speed of the camera was gradually increased and the tracking performance of each method is evaluated. Experiments were conducted at different lighting conditions to assess each method's robustness. For intensity image-based detection, the well-known Kanade-Lucas-Tomase (KLT) \cite{klt1991} tracker was coupled with conventional CHT to track the detected circular holes. As expected, intensity image-based tracking deteriorates at higher speeds due to motion blur; and entirely fails to detect the hole at low-light due to the inclarity of the image despite the increased exposure time. Neuromorphic vision-based perception on the other hand remains persistent despite these variations.

\subsection{Nutplate holes drilling performance}
\label{sec_exp_drill}

\begin{figure}[]
 \centering
      \includegraphics[width=\linewidth]{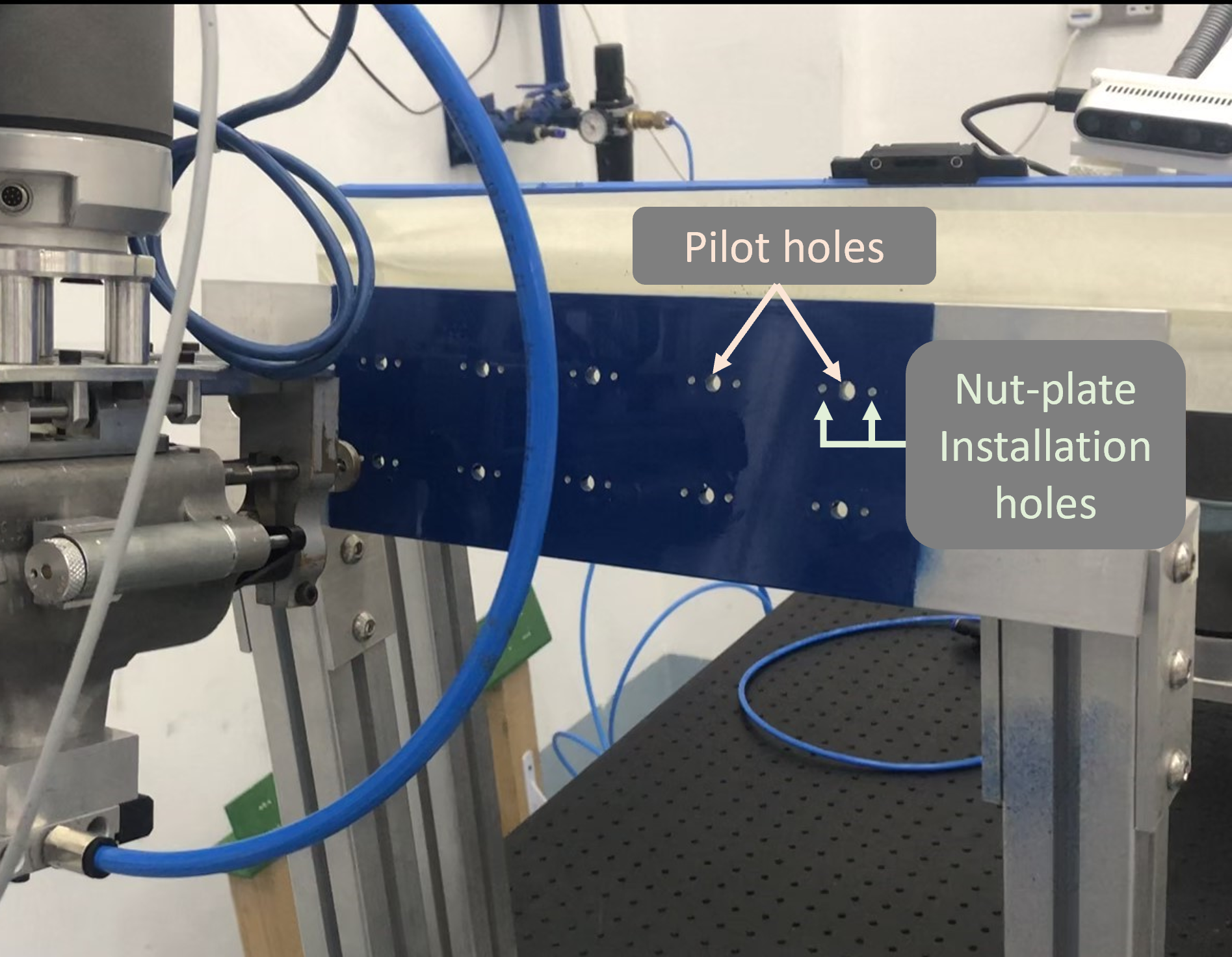}
\caption{The target workpiece after drilling the nutplate installation holes using the proposed neuromorphic vision based robotic drilling approach.}
\label{fig_drilled_plate}
\vspace{-1em}
\end{figure}

The performance of the overall nutplate holes' drilling process is presented in this section. Tests were conducted using the setup shown in Figure \ref{fig_robot_exp_setup} with five workpieces placed differently in the environment. The mobile robot autonomously navigates to the front of each of the workpieces, and then our novel neuromorphic visual control pipeline described in sections \ref{sec_multi_view}, \ref{sec_hole_det}, and \ref{sec_control} control the manipulators to perform the drilling objective. A video demonstration of these experiments can be viewed through this link: \url{https://drive.google.com/file/d/1q9QwPvkd7ZcEBcGMIxIVy2r_iRfTKCSe/view?usp=sharing} \cite{paper_video}.  We assess the drilling performance in terms of the positional error of the nutplate holes. Table \ref{tab_drilling_pos_result} presents the per-hole positional error across the five different workpieces and Figure \ref{fig_drilled_plate} shows an example workpiece after drilling the nutplate holes.


\renewcommand{\arraystretch}{1.5}
\begin{table*}[]
\caption{Position error in mm for the nutplate holes' drilling experiments across multiple workpieces.}
\centering
\begin{tabular}{|l|c|c|c|c|c|c|c|c|c|c|c|c|c|c|c|}
\hline
\textbf{Workpiece} & \multicolumn{10}{c|}{\textbf{Hole ID}} & \multirow{2}{3em}{\textbf{Mean}}  & \multirow{2}{3em}{\textbf{Max}} & \multirow{2}{5em}{\textbf{Standard Deviation}} \\ \cline{2-11}
& \textbf{1} & \textbf{2} & \textbf{3} & \textbf{4} & \textbf{5} & \textbf{6} & \textbf{7} & \textbf{8} & \textbf{9} & \textbf{10} & & & \\ \hline

\textbf{A} & 0.039 &   0.134 &   0.157  &  0.100 &   0.131   & 0.075  &  0.094  &  0.052 & 0.077  &  0.103 & 0.096 & 0.157 & 0.037 \\ \hline

\textbf{B} & 0.141 &   0.102 &   0.064   & 0.062 &   0.110 &   0.068  &  0.076  &  0.099 &   0.070 &   0.063 & 0.085 & 0.141 & 0.026 \\ \hline

\textbf{C} & 0.069  &  0.034  &  0.101 &   0.030 &   0.090 &   0.114 &   0.082 &   0.113  &  0.052  &  0.023 & 0.071 & 0.114 & 0.035 \\ \hline

\textbf{D} & 0.183 &   0.003  &  0.085  &  0.100 &   0.175  &  0.175 &   0.047  &  0.061  &  0.017  &  0.098 & 0.095 & 0.183 & 0.066 \\ \hline

\textbf{E} & 0.011  &  0.086  &  0.106  &  0.108 &   0.124  &  0.077 &   0.066  &  0.144  &  0.126  &  0.051 & 0.091 & 0.145 & 0.040 \\ \hline

\multicolumn{11}{r|}{\textbf{Aggregate}} & \bf{0.088} & \bf{0.183} & \bf{0.042} \\ \cline{12-14}

\end{tabular}
\label{tab_drilling_pos_result}
\end{table*}







Quantitative results show that our proposed neuromorphic vision-based approach is capable of precisely drilling nutplate holes with an average error of less than 0.1 mm. These results conform with the precision requirements of a large variety of processes in the automotive and aerospace manufacturing industries. This validates the use of neuromorphic vision for precise manufacturing tasks and highlight the potential of using neuromorphic cameras for faster and more reliable automated manufacturing. The obtained results also prove the effectiveness of our proposed algorithms in employing the advantages of neuromorphic cameras while addressing several of their challenges in terms of unconventional data output, and relatively low resolution. We would like to indicate that the nature of the performed drilling process, which includes inserting the clamp mandrel in a pilot hole, contributes to further minimizing the positional errors. During this peg-in-hole stage, the end-effector's pose can be driven to better alignment with the reference hole due to the compliance of the manipulator, which can further reduce any errors resulting from the visual guidance process.

\section{Conclusions and Future Work}

In this paper, we presented the first system that employs the recent neuromorphic vision technology for robotic machining applications. In particular, we have developed a complete visual guidance solution that precisly positions the robot relative to the desired workpiece with sub-millimeter accuracy using two consequent stages of perception and control. The first stage utilizes a multi-view 3D reconstruction approach and PBVS for the initial alignment of the robot's end-effector. Concurrently, the second stage regulates any residual errors using a novel event-based hole detection algorithm and IBVS.

We have validated our system experimentally for a nutplate hole drilling application using a collaborative robot manipulator, an iniVation neuromorphic camera, and a customized end-effector. Our quantitative results show  that the presented neuromorphic vision-based solution can successfully drill the target holes with an average positional error of less than 0.1mm. Our tests also verify that the use of neuromorphic cameras overcomes the lighting, speeds and motion blur challenges associated with the use of conventional frame-based cameras. These results demonstrate the potential of using neuromorphic cameras in precise manufacturing processes, where they can facilitate faster and more reliable production lines. 

For future work, we aim to improve the normal adjustment and orientation control aspects of our robotic drilling system. In our current system, the only measurement on workpiece orientation is obtained using the multi-view reconstruction step; and any orientation errors resulting from this step would not be corrected for, unlike position measurements which are further refined using circular hole detection. Although the two-sided clamping and the compliance of the collaborative robot can passively drive the end-effector towards better normality with the workpiece, a more precise and reliable normal alignment method is required to expand the range of manufacturing processes our system can perform. To this end, We will investigate the application of visual tactile sensing \cite{baghaei2020neuromorphic, 1_baghaei2020neuromorphic} for the normality control in robotic machining processes. 







\printcredits

\bibliographystyle{elsarticle-num}

\bibliography{main}

\end{document}